\documentclass[11pt, a4paper]{article}
\usepackage{fullpage}

\newif\ifabstr
\abstrfalse

\usepackage[utf8]{inputenc}
\usepackage[numbers]{natbib}
\usepackage{booktabs}
\usepackage{multirow}
\usepackage[tableposition=top,font=small,labelfont=bf]{caption}
\usepackage[section]{placeins}
\makeatletter
\g@addto@macro\@floatboxreset{\def\@tempa{table}\ifx\@captype\@tempa\small\fi}
\makeatother

\usepackage{amsmath}
\usepackage{amssymb,amsthm,graphicx}
\usepackage[dvips]{epsfig}
\usepackage{amsfonts}
\usepackage{xcolor}
\usepackage{microtype}
\usepackage{xurl}
\usepackage{hyperref}
\usepackage{enumerate}

\title{Machine learning approaches to seismic event classification in the Ostrava region}

\author{
Michael Skotnica$^1$,
Marek Pecha$^{1}$,
Jana Ru\v{s}ajov\'{a}$^1$,
Bohdan Rieznikov$^{1,2}$,\\
V\'{i}t Wandrol$^{1,3}$,
Mark\'{e}ta R\"{o}snerov\'{a}$^1$,
Jarom\'{i}r Knejzl\'{i}k$^1$ \\[2mm]
$^1$ Institute of Geonics of the Czech Academy of Sciences, Ostrava, Czechia \\[2mm]
$^2$ Unicorn University, Prague, Czechia \\[2mm]
$^3$ Brno University of Technology, Czechia \\[2mm]
}

\date{}

\begin{document}

\maketitle

\begin{abstract}
The north-eastern part of the Czech Republic, including the Czech part of the Upper Silesian Coal Basin,
is among the most seismically active areas of the country.
The seismic stations in the region record mining-induced events,
which continue after the cessation of mining, natural earthquakes,
which are rare but may originate in the same area, and quarry blasts.
A rapid and reliable distinction between the three types is required
for the routine evaluation of the continuously recorded data.
We study their automatic discrimination with neural networks from short three-component windows
starting one second before the P-wave onset.
The main data set is the archive of the Seismic Polygon Fren\v{s}t\'{a}t (1992--2002),
a triggered network whose records were recovered from the original format,
cleaned with deterministic and learned quality checks and completed with automatically picked P-wave arrivals;
it contains about 40,000 records of all three event types. Among five compared architectures,
convolutional networks operating on short-time spectrograms outperform recurrent networks
operating on magnitude spectra or on the raw waveform.
The best model, EfficientNet-B0 applied to 24-s windows, reaches a macro-averaged F1 score of 0.96,
and blasts and natural earthquakes are practically never confused with each other;
class weighting and undersampling, the usual remedies for the strong class imbalance, lower the score.
Finally, we examine the transfer to the station OKC, which monitors the region today with different instruments.
A model trained on the archive alone fails on the modern records,
whereas a single model trained on both data sets reaches a macro-F1 score of 0.96 on the archive
and 0.92 on the modern station,
where the accuracy is limited by the small number of catalogued natural earthquakes
rather than by the model.
\end{abstract}

\maketitle
\section{Introduction}
The northeastern region of the Czech Republic (Jesen\'{i}ky Mountains, Opava Region, Ostrava Region, Upper Morava Valley) 
is one of the most seismically active areas in the country.
In addition to tremors induced by hard coal mining in the Upper Silesian Coal Basin, natural earthquakes also occur here.
Such earthquakes are mostly very weak with a magnitude of less than 0 and are only recorded by instruments.
Occasionally, there were stronger earthquakes felt by people
(e.g., Jesen\'{i}ky Mountains in 1935, 1986, and 2012, or Opava region in 1931, 1936, and a swarm in 1993).
Mining in the Czech part of the basin ended in 2026, but it continues on the Polish side, and induced events are still recorded in the Czech part after the end of mining (e.g., due to the flooding of closed mines).
In addition to mining, several active quarries operate within the study area (e.g., Bohu\v{c}ovice and Jakub\v{c}ovice -- greywacke; \v{S}tramberk -- limestone). Quarry blasting is routinely detected by the regional seismic stations and contributes to the overall seismic event catalogue.
Between 1992 and 2002 the area was monitored by the Seismic Polygon Fren\v{s}t\'{a}t (SPF), a triggered network whose archive is the main data source of this study (Section~\ref{sec:spf}); since 2007 digital records are available from the station Ostrava-Kr\'{a}sn\'{e} Pole (OKC) of the Czech Regional Seismic Network (Section~\ref{sec:okc}).

Since approximately 2010, the stations of the Institute of Geonics (IGN) have switched to continuous recording with a sampling frequency of 100\,Hz,
which has resulted in a substantial increase in the amount of data processed.
An automated classification of the seismic events in this area is therefore necessary:
induced events, which continue to occur after the end of mining,
must be quickly distinguished from natural tectonic earthquakes and from quarry blasts.
Machine learning offers a natural solution to this task.

In this study, we apply several modern machine learning methods trained on the SPF and OKC data to classify seismic events into three categories:
\begin{itemize}
    \item Mining-induced seismic events (M),
    \item Natural tectonic seismic events (T),
    \item Surface explosions, mainly quarry blasts (E).
\end{itemize}
This is the distinction made in the routine seismological service in the region, and it is also the practically relevant one: induced events and natural earthquakes may originate in the same area (Section~\ref{sec:locality}), while quarry blasts are a frequent third source that must not be confused with either.

The discrimination of anthropogenic from natural seismicity is a long-standing problem~\citep{WiemerBaer2000, Dahm2013, Grigoli2017}, and machine learning has become the standard tool for it in the last decade~\citep{Kong2019, Mousavi2022}. Deep networks have been used to separate quarry blasts from earthquakes in regional networks~\citep{Linville2019, Renouard2021} and, closer to the present setting, to classify microseismic events and blasts recorded in mines~\citep{Vallejos2013, Dong2016, Bi2021, Dong2023}. Most of these studies, however, deal with a binary task and with a single, homogeneous data set. In our previous work we also addressed binary problems: the separation of quarry blasts from tectonic events on the OKC station and a first attempt at tectonic versus mining-induced events on a small sample of the SPF data~\citep{Skotnica2025medgu}, and the separation of quarry blasts from mining-induced events on the stations MORC and VRAC of the Czech Regional Seismic Network~\citep{Skotnica2026egu}.

The present paper differs in two respects. First, it addresses the full ternary problem. This is only possible because the SPF archive is, to our knowledge, the only data set in the region in which all three event types are available in sufficient numbers; the catalogue of the current station OKC used here contains only about 150 natural earthquakes, fewer than twenty of them from the Ostrava area itself. Second, the region is monitored today by a different generation of instruments than in the 1990s. A model that is to run on the present network must therefore either transfer from the archive or be trained on both generations of data, and we examine what this requires.

Although the data come from one region, neither the data preparation
pipeline nor the classifiers contain anything specific to it; both can be
applied to other seismically monitored mining areas after retraining on
local records. The Ostrava region is, however, a particularly suitable
test case, because all three event types occur there within a small
area and are recorded by the same stations.

The contributions of this paper are the following.
\begin{itemize}
    \item We recover the complete SPF archive (1992--2002) from its original triggered-system format, clean it with a combination of deterministic and learned quality checks, and pick the P-wave arrivals, which yields a labelled set of about 40,000 three-component records of the three event types (Sections~\ref{sec:input-data}--\ref{sec:cleaning}).
    \item We compare five neural architectures for the ternary classification of short windows starting one second before the P-wave onset, including an analysis of the record length, of the time--frequency representation, and of the treatment of the strong class imbalance; the best model reaches a macro-averaged F1 score of 0.96 (Section~\ref{sec:results}).
    \item We study the transfer of these models to the present-day station OKC: a model trained on the archive alone does not transfer across the two generations of instruments, whereas a single model trained on both data sets performs well on both, and the number of catalogued natural earthquakes on the modern station, not the model, turns out to be the limiting factor (Section~\ref{sec:transfer}).
\end{itemize}

\section{Study area and seismic monitoring}
\label{sec:locality}

\subsection{Study area}
The study locality of Silesia and Northeast Moravia can be characterized by low natural seismic activity based on 30 years of seismological observations realized in the territory~\cite{Prachar2022}.
There are several seismically active areas, some of them are characterized even by occurring of seismic swarms (e.g., Opava in 1931 and Hradec nad Moravicí in 1993).
The strongest instrumentally recorded earthquake in this locality, local magnitude 3.5,
occurred in 2017 near the Hlu\v{c}\'{i}n town~\cite{SilenyZednik2018}.
Nevertheless, stronger historical earthquakes with intensities up to $I_0=7$ are documented at the locality (e.g. Ostrava in 1786 $I_0=7$; Bohum\'{i}n 1259 $I_0=6$; Opava 1931 $I_0=6$~\cite{Dopita1997}); see Figure~\ref{f:map}.
Mining activities related to the underground exploitation of hard coal from Czech and Polish mines at the territory of the Upper Silesian Coal Basin (USCB) represent another important source of seismic activity in the region.
It can be seen from Figure~\ref{f:map}, that the strongest natural earthquakes (2017, 1786, 1259) are located close to mining claims in the Czech part of USCB.
Thus, it could be expected that both types of seismic events, natural and mining-induced (induced by mining or by flooding of closed parts of mines),
could originate at the same locality.
Therefore, investigation and recommendation of suitable discrimination techniques for determination of event's origin should be useful for future seismic observations at the territory.

The Czech part of the USCB is situated on the northeastern edge of the Czech Republic and geologically it belongs to the segment of Variscide orogeny. Upper Carboniferous strata represent hard rocks of basin including coal seams.
Carboniferous formations are buried with Quaternary and Tertiary sedimentary rocks.
The thickness of these sedimentary rocks is variable -- from first meters to several hundreds of meters.
For a detailed description of the tectonic structure and stress conditions in the USCB, see~\cite{Waclawik13, Grygar2011, Ptacek2012}.
The territory of SPF seismic network is characterized by close contact between the Outer Carpathians and the Bohemian Massif.
The margin of the Outer Carpathians overfault to the Miocene foredeep follows the southern margin of the Ostrava Quaternary basin~\cite{Holub04}.

\subsection{The SPF network (1992--2002)}
\label{sec:spf}
\begin{figure}
	\centering
		\includegraphics[scale=1.5]{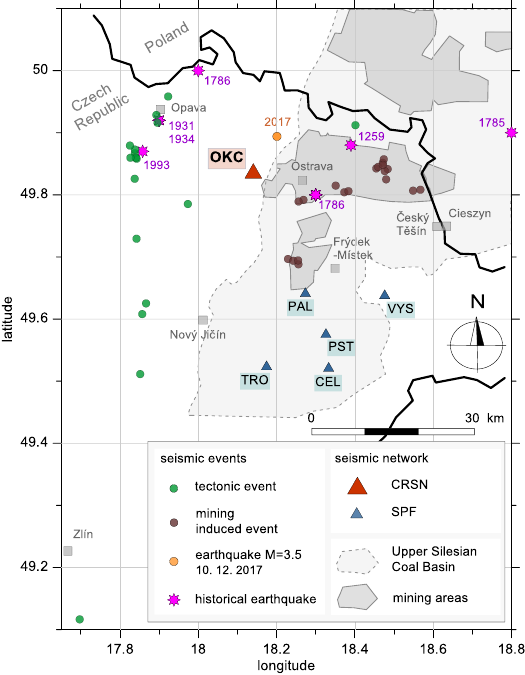}
	\caption{A map of the strongest earthquakes in the Northeastern region of the Czech Republic, the stations of the Seismic Polygon Fren\v{s}t\'{a}t (SPF), the Czech part of The Upper Silesian Coal Basin (light grey).}\label{f:map}
\end{figure}

The SPF seismic network consisted of 5 seismic stations located at the territory of $20 \times 20$ km in the south part of the USCB.
See Figure~\ref{f:map}. Coordinates and names of stations are presented in Table~\ref{t:SFP_coordinates}. Seismic stations were equipped with 3-component seismometers WDS-202 ($f_0 = 2\,\mathrm{Hz}$) placed in a shallow boreholes at the depth approximately 30~m below the surface. Seismic signal was recorded by the apparatus PCM3-T in triggered regime: recording was started by an STA/LTA trigger~\cite{Allen1978}, i.e.\ when the ratio of the short-term to the long-term average of the signal amplitude exceeded the trigger level at least at one station. The sampling frequency of recorded signal was 125\,Hz. The data recorded from all stations was concentrated by the relay station situated at the top of the Beskydy mountains in a communication tower equipped with a data concentrator based on the SAPI-1 microcomputer. Afterwards data were transmitted by radio transmitter to the recording center in Ostrava. Recorded data files were converted into the ESSTF/2 format, which was modified from the standard ESSTF format by transferring 4 byte signal amplitude information into 2 byte form. It enabled a reduction of the recorded data volume by half. A detailed description of the network is given in~\cite{Holub04SPF}.
The special seismic interpretation system WAVE~\cite{Toth1992} was developed for recorded digital data processing
with individual program packages enabling given analysis -- localization, spectral analysis, polarization analysis, focal mechanism determination.

During the entire period, the SPF recorded approximately 17,384 events,
mostly induced seismic events related to mining in the Ostrava-Karvin\'a mining district (OKR) and Poland,
a total of 14,144, followed by 2,516 natural earthquakes.
Another type of phenomenon was quarry blast. There were 639 such events recorded.
In addition, 85 other phenomena were not classified. 
Of the 14,144 induced events, 5,841 originated in the Czech part of the basin, the OKR; the remaining ones came from the Polish part of the USCB and, a minority, from the vicinity of Ojc\'{o}w (Poland).
Seismic energy was determined only for the events from the Czech part: 195 events had an energy of the order of $10^{2}$~J, 3,586 of $10^{3}$~J, 1,852 of $10^{4}$~J,
182 of $10^{5}$~J, 23 of $10^{6}$~J, and 3 of $10^{7}$~J.

For the purpose of this study, data set of seismic events recorded by the SPF network (location of stations in Figure~\ref{f:map}) during the period 1992 -- 2002 was compiled.
The SPF network was in operation 10 years with no interruption and no change of seismic instrumentation, so the recorded data are suitable for the purpose of this study, it means investigation of discrimination techniques to distinguish natural and mining-induced seismic events. Epicenters of selected seismic events recorded during 10 years of monitoring at the stations of SPF cover area of $60 \times 60$ km and they are displayed on the map in Figure~\ref{f:map}. Tectonic events are located mainly north-westerly and westerly of seismic network, only one event is located close to mining claim in the Czech part of USCB. Mining-induced events cover all undermined areas in the Czech part of USCB (Ostrava sub-basin, Pet\v{r}vald sub-basin, Karvin\'{a} sub-basin, Sta\v{r}\'{i}\v{c} area).

\begin{table}
\centering
\caption{The coordinates of the stations from SPF.}\label{t:SFP_coordinates}
\begin{tabular}{llccc}
\toprule
Code & Name & Lon ($^\circ$E) & Lat ($^\circ$N) & h (m) \\
\midrule
TRO & Trojanovice & 18.174 & 49.523 & 435 \\
PST & Pstru\v{z}\'{i} & 18.326 & 49.575 & 461 \\
PAL & Palkovick\'{e} H\r{u}rky & 18.273 & 49.640 & 517 \\
CEL & \v{C}eladn\'{a} & 18.333 & 49.520 & 454 \\
VYS & Vy\v{s}n\'{i} Lhoty & 18.476 & 49.637 & 456 \\
\bottomrule
\end{tabular}
\end{table}

\subsection{The station Ostrava-Kr\'{a}sn\'{e} Pole (2007--present)}
\label{sec:okc}
The station Ostrava-Kr\'{a}sn\'{e} Pole (OKC) --- $49.83460^\circ$N, $18.13990^\circ$E, $h = 250$\,m --- was established in 1983.
It is operated by the Institute of Geonics of the Czech Academy of Sciences (CAS), the Institute of Geophysics of the CAS
and the V\v{S}B -- Technical University of Ostrava.
It is part of the Czech Regional Seismic Network as well as the international network of stations.
It is equipped with three-component short-period SM3 sensors
--- suitable mainly for registering nearby induced seismic phenomena in the USCB ---
and the broadband seismometer G\"{u}ralp CMG-3ESP for registering more distant earthquakes;
the records of the latter are used in this study.
The sensors are located in a shallow tunnel with a small seismic disturbance.
The Quanterra Q330S registration apparatus is located in the building of the Astronomical Observatory of Johann Palisa.
The station operates online and digital data (since 2007) are transferred in real time to the Institute of Geophysics (IG).
It mainly contributes to the rapid localization of mining-induced seismic events, local tectonic phenomena and major world earthquakes.

The OKC station is also part of a local seismic network that monitors earthquakes
associated with coal mining in the USCB (Czech Republic and Poland) and natural seismicity in Northern Moravia and Silesia.

\section{Seismic input data}
\label{sec:input-data}

Since there is not enough local natural seismic events in the OKC dataset recorded in 2007-- present,
we also incorporated the data from SPF, which captured the tectonic events occurring in 1990s,
providing a valuable dataset for training our model.
The SPF archive is therefore the main training set of this study, and the OKC data serve as the target of the transfer experiments in Section~\ref{sec:transfer}.

\subsection{SPF archive: format conversion}
The data of SPF was originally stored in a custom ESSTF/2 format
based on ESSTF format which was a European format for seismic data by Lennartz. In 80s and 90s it was a standard format for data produced by systems in triggered regime~\cite{Raykova00, Holub04SPF}.
Therefore, we implemented a decoder to the miniSEED format which is a modern standard for exchange of earthquake data~\cite{MSEED24}.

\subsection{OKC data set}
\label{sec:okc-data}
The second data set consists of events recorded by the station OKC
(Section~\ref{sec:okc}). The annotations are taken from the monthly
seismic bulletins of the IGN for this station, compiled
by one of the authors (J.~Ru\v{s}ajov\'{a}) and available
online\footnote{\url{https://www.ugn.cas.cz/?p=bulletin}}. The bulletins
cover the period from January 2007 to November 2023 (203 monthly
bulletins); from each listed event we took the manually picked arrival
times of the Pg and Sg phases and the short description of the source,
from which the class label (E, M or T) was derived: blasts are marked
``EXP'', optionally with the name of the quarry; mining-induced events
carry the name of the mining area (the Czech part of the basin, Polish
mining districts, Sta\v{r}\'{i}\v{c}); natural earthquakes are marked
``tectonic event'' or with the source region (e.g.\ Slovakia, West
Bohemia, High Tatras).

The waveforms are the three components of the broadband Güralp CMG-3ESP
sensor of the station, sampled at 100\,Hz. The recording system changed
once during the period: an Earth Data PR6-24 digitizer was in use until
23~March 2011 and a Quanterra Q330S afterwards. For each event a window
was cut from the digital record of the station starting 1~s before the Pg arrival,
consistently with the SPF records, and ending 20~s after the Sg arrival.
For the 219 events without a picked Sg arrival the window
ends 30~s after Pg. The resulting records are between 22 and 79~s long
(median 25~s), i.e.\ comparable with the triggered SPF records.

In total, the OKC data set comprises 20,410 events: 17,928 mining-induced
events, 2,329 quarry blasts and 153 natural earthquakes. The natural
earthquakes --- the class of greatest interest for the monitoring of the
region --- are thus even rarer than in the SPF archive (0.75\,\% against
4.7\,\% of the records) and too few for a reliable classifier of this
class on this data set alone, which motivates the transfer experiments
in Section~\ref{sec:transfer}.
The natural earthquakes of the two data sets also differ in composition.
The SPF network was triggered by every event and recorded the
considerable local tectonic activity of the 1990s --- more than sixty
earthquakes near Hradec nad Moravic\'{i} and Opava, which account for
several hundred records of the tectonic class --- together with
earthquakes from more distant parts of the region (e.g.\ the
Kr\'{a}lick\'{y} Sn\v{e}\v{z}n\'{i}k massif) and beyond. The OKC data set
is limited to the events listed in the bulletins of the IGN, which
cover the seismicity of the region and, occasionally, of
its surroundings (Slovakia, Poland, Austria, West Bohemia); fewer than
twenty of its 153 natural earthquakes originate in the Ostrava area
itself, so that the local earthquakes, well represented in the archive,
are nearly absent from the modern data set. Distant earthquakes, which
the broadband station records as well, are evaluated by the IG and do
not appear in these bulletins. The complete
catalogues of the Czech Regional Seismic Network would enlarge the
tectonic class considerably; they have not been used in this study,
whose primary subject is the SPF archive and the routine processing of
the regional seismicity at the IGN, and their
inclusion is the obvious next step for the present-day network
(Section~\ref{sec:discussion}).
The records are stored in raw digital counts; since every record is
normalized per event before classification (Section~\ref{sec:preprocessing}),
no instrument correction is applied.

\section{Data cleaning and preparation}
\label{sec:cleaning}
As we described above, the analyzed data set consists of seismic signals
recorded in a triggered regime at 5 stations of SPF network between the years 1992 -- 2002.
Although all seismic events were always recorded at all five stations,
not all records were suitable for analysis due to the poor quality of the signal, which could be distorted. Moreover, sometimes there is missing P- or S-wave arrival.

For 4,568 records, the P-wave arrival times were picked manually by one of the authors (J.~Ru\v{s}ajov\'{a}) using the WAVE software (Section~\ref{sec:spf}).

In this section, we describe how we prepare the SPF data in the format compatible with our modern approach used on the modern data from the OKC station. That is, each event was recorded in a separate miniSEED file containing all three signal components and with a known time of P-wave arrival. The OKC records (Section~\ref{sec:okc-data}) are cut from continuous data at the bulletin picks and did not require this cleaning.

\subsection{Pipeline Overview}

The cleaning pipeline proceeds as follows:

\begin{enumerate}
  \item \textbf{Autoencoder training.} If a list of known-good
        (``OK'') records is provided (the records with manually picked P-wave arrivals, see above), a convolutional denoising autoencoder is
        trained on sliding windows extracted from these records
        (Section~\ref{sec:learned}). An anomaly threshold is calibrated on a
        held-out validation split.
  \item \textbf{Per-file testing.} Every file in the input list is read,
        its channels are sorted, and the following tests are applied:
        \begin{enumerate}
          \item deterministic single-channel and cross-channel checks (Section~\ref{sec:deterministic}),
          \item autoencoder anomaly scoring, if a trained model is available
                (Section~\ref{sec:learned}).
        \end{enumerate}
        In parallel, a spectral feature vector is computed for each record with
        finite (non-NaN) features (Section~\ref{sec:learned}); records
        containing NaN values are flagged directly.
  \item \textbf{$k$-means outlier detection.} After all files are processed,
        $k$-means clustering~\cite{macqueen67} with $k=2$ is fitted on the collected feature
        vectors, and the minority cluster is marked as corrupted
        (Section~\ref{sec:learned}).
  \item \textbf{P-wave arrival picking using PhaseNet} (Section~\ref{sec:picking}).
  \item If a record or its channel is marked as corrupted using the procedures above, or PhaseNet was not able to find the P-wave arrival time, it is removed.
\end{enumerate}

\subsection{Deterministic signal-quality checks}
\label{sec:deterministic}

\begin{figure}
    \centering
    \includegraphics[width=0.9\linewidth]{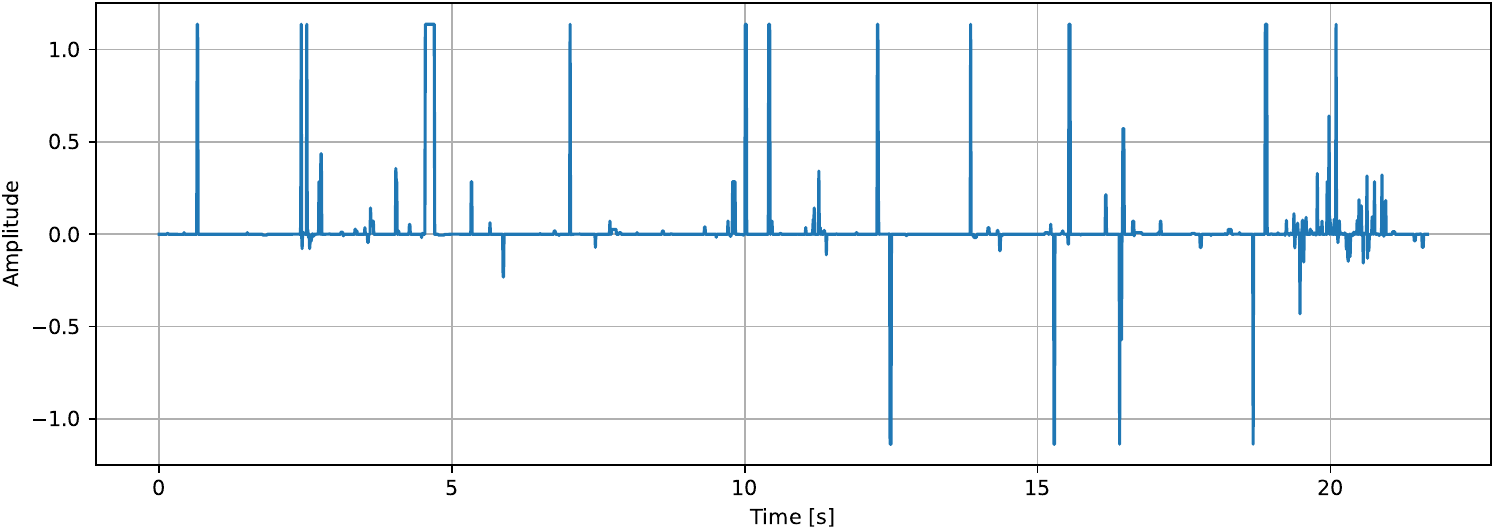}\\[-1ex] {\small (a)}\\[1ex]
    \includegraphics[width=0.9\linewidth]{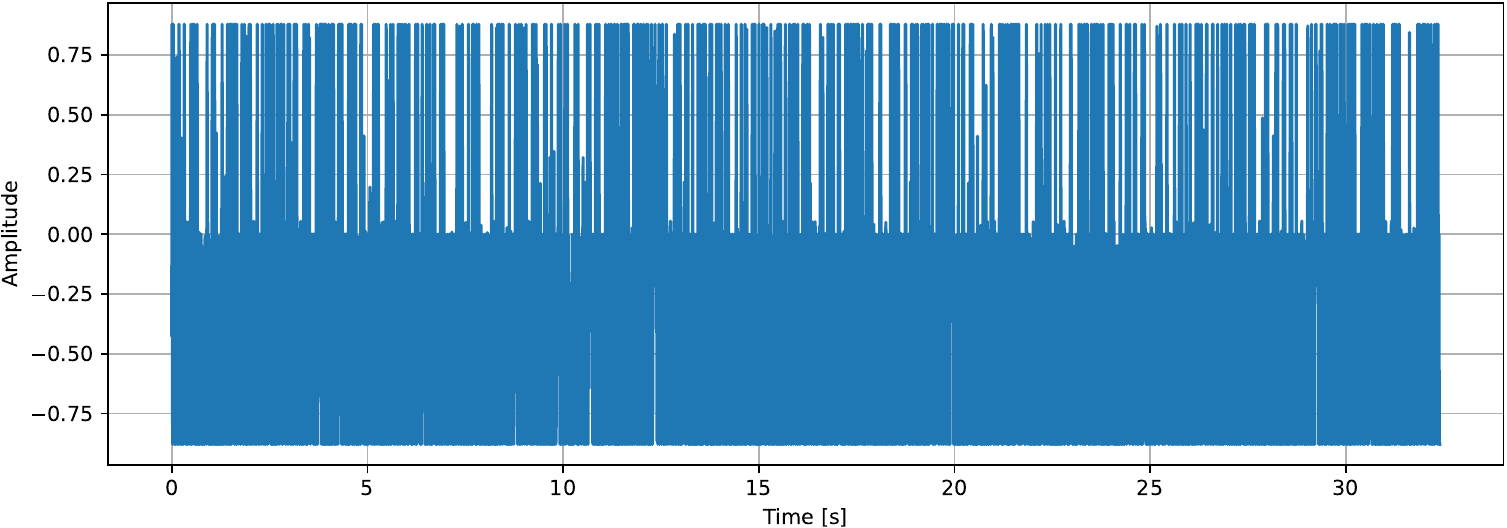}\\[-1ex] {\small (b)}\\[1ex]
    \includegraphics[width=0.9\linewidth]{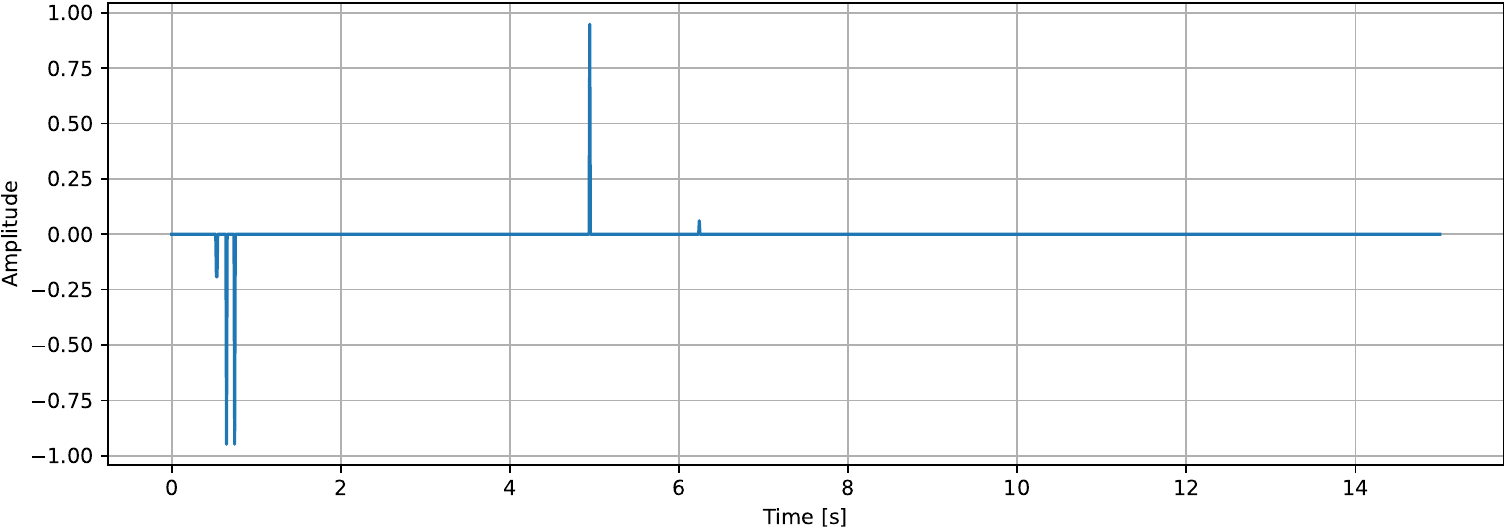}\\[-1ex] {\small (c)}
    \caption{Examples of corrupted SPF records detected by the deterministic
    checks: (a) a clipped signal (amplitude threshold), (b) a distorted signal
    with a prolonged high amplitude (near-saturation ratio), (c) a very weak
    signal with several unrealistic spikes (mean-concentration check).}
    \label{f:corrupted}
\end{figure}

The deterministic tests are simple, interpretable rules with tunable
thresholds; a record is flagged if any test fires on any channel. The
parameter values are listed in Table~\ref{tab:params}, and
Figure~\ref{f:corrupted} shows typical records caught by them. The first
five tests are applied independently to every channel $x$ of length $T$
of the record, the last two compare the three channels with each other.

\paragraph{Amplitude threshold.}
The maximum absolute amplitude must stay below a threshold $A$, i.e.\
$\max_i |x_i| < A$ (in normalized amplitude units). This targets saturated
or clipped traces (Figure~\ref{f:corrupted}a).

\paragraph{Near-saturation ratio.}
The fraction of samples close to the amplitude threshold must be small:
\begin{equation}
  \frac{1}{T}\,\bigl|\{\, i : |x_i| \ge \mu_A A \,\}\bigr| < \rho_A ,
\end{equation}
where $\mu_A$ is the threshold multiplier and $\rho_A$ the maximum allowed
ratio. A large fraction of near-maximal samples indicates prolonged clipping
rather than a genuine transient (Figure~\ref{f:corrupted}b).

\paragraph{Mean-concentration check.}
Let $m = \frac{1}{T}\sum_i |x_i|$ be the mean absolute amplitude; we require
\begin{equation}
  \frac{1}{T}\,\bigl|\{\, i : |x_i| \le m \,\}\bigr| < \rho_m .
\end{equation}
If almost all samples lie below the mean, the amplitude distribution is
dominated by a few extreme outliers, e.g.\ isolated spikes on an otherwise
flat trace (Figure~\ref{f:corrupted}c).

\paragraph{Zero-ratio check.}
The fraction of exactly-zero samples must be small,
$\frac{1}{T}\,|\{\, i : x_i = 0 \,\}| < \rho_0$; this detects gaps, dead
channels, and zero-filled segments.

\paragraph{Max-to-mean ratio.}
The ratio of the maximum to the mean absolute amplitude must be bounded,
$\max_i |x_i| \,/\, m \leq \kappa$; an extremely large ratio indicates
isolated spikes inconsistent with a physical seismic signal.

\paragraph{Inter-channel amplitude ratio.}
Let $M_Z, M_N, M_E$ denote the maximum absolute amplitudes of the three
channels. The record is flagged if any pairwise ratio $M_Z/M_N$, $M_Z/M_E$,
$M_N/M_E$ leaves the band $[1/R,\; R]$. A channel that is an order of
magnitude stronger or weaker than its siblings typically indicates a gain
error or a partially dead component.

\paragraph{Spectral correlation.}
Each channel is centred by subtracting its median and its amplitude spectrum
$\hat{x} = |\mathrm{FFT}(x - \operatorname{med} x)|$, computed with the fast
Fourier transform (FFT), is normalized to unit
Euclidean norm, $\bar{x} = \hat{x} / \lVert \hat{x} \rVert_2$. The similarity
of two channels is the inner product of their normalized spectra, e.g.\
$c_{ZN} = \langle \bar{Z}, \bar{N} \rangle \in [0, 1]$, and the record is
flagged if $\min( c_{ZN}, c_{ZE}, c_{NE} ) < c_{\min}$. Genuine seismic
events excite all three components with similar spectral content, whereas
instrumental artifacts usually affect a single channel.

\begin{table}[htbp]
\centering
\caption{Parameters of the deterministic checks.}
\label{tab:params}
\begin{tabular}{lll}
\toprule
Symbol  & Value & Meaning \\
\midrule
$A$                  & $1.0$    & maximum allowed amplitude \\
$\mu_A$     & $0.8$    & near-saturation level (fraction of $A$) \\
$\rho_A$       & $0.35$   & max.\ ratio of near-saturated samples \\
$\rho_m$             & $0.95$   & max.\ ratio of samples below the mean \\
$\rho_0$             & $0.4$    & max.\ ratio of zero samples \\
$\kappa$           & $10^{5}$ & max.\ max-to-mean amplitude ratio \\
$R$                     & $10$     & max.\ inter-channel amplitude ratio \\
$c_{\min}$         & $0.6$    & min.\ spectral similarity \\
\bottomrule
\end{tabular}
\end{table}

\subsection{Learned checks: spectral clustering and autoencoder}
\label{sec:learned}

\paragraph{Spectral clustering.}
For the unsupervised detection, every record is mapped to a fixed-length
feature vector. Each channel $x$ is first z-score normalized,
\begin{equation}
  \tilde{x}[t] = \frac{x[t] - \mu_x}{\sigma_x},
  \label{eq:zscore}
\end{equation}
where $\mu_x$ and $\sigma_x$ are the mean and the standard deviation of
the channel over the record; this normalization removes the amplitude
scale and the offset of the record and is applied to all model inputs in
this paper. The normalized channel is then transformed to its
log-magnitude spectrum, i.e.\ the representation of Eq.~(\ref{eq:fft}) in
Section~\ref{sec:tf-repr} computed with $N = 4096$ samples (longer signals
are truncated, shorter ones zero-padded) and compressed by
$\log(1 + |X|)$; the three per-channel spectra are concatenated. Records for
which the features contain NaN values (e.g.\ due to corrupted samples) are
flagged directly and excluded from clustering. The feature vectors of all
valid records are clustered with $k$-means~\cite{macqueen67} ($k = 2$, 20 restarts, fixed
random seed). Under the assumption that corrupted records are a minority and
that their spectra differ systematically from those of clean records, the
smaller of the two clusters is interpreted as the corrupted class. Note that
this method is purely relative: it only separates records that are
spectrally atypical within the given batch, and it presumes that anomalies
form the minority cluster which is our case.

\paragraph{Autoencoder.}
The third detector is a one-dimensional convolutional denoising autoencoder
operating on short sliding windows of the three-component signal. Each
record is padded or cropped to 8,192 samples, z-score normalized per
channel, and cut into overlapping windows of $T_w = 256$ samples with hop
64. The encoder consists of three convolutional layers, the latter two with
stride 2, compressing the window to a latent representation at one quarter
of the temporal resolution; the decoder mirrors this structure with two
transposed convolutions and a final convolutional projection back to three
channels (Table~\ref{tab:ae}). The autoencoder is trained \emph{only on
windows from known-good records}: the training input is corrupted by
additive Gaussian noise and random feature dropout, while the reconstruction
target remains the clean window. The anomaly score of a window is its mean
absolute reconstruction error,
\begin{equation}
  s(w) = \frac{1}{3\,T_w} \sum_{t=1}^{T_w} \sum_{c=1}^{3}
         \bigl| \hat{w}_{t,c} - w_{t,c} \bigr|,
\end{equation}
where $\hat{w}$ is the reconstruction of window $w$. Because the model is
trained on clean data only, windows containing artifacts it has never seen
are reconstructed poorly and receive high scores. The per-window scores of a
record are aggregated by their maximum, and the decision threshold is set to
the 99th percentile of the scores on a held-out validation split (20\,\%) of
the clean training windows, so that by construction about 1\,\% of clean
windows would be falsely flagged. A record whose aggregated score exceeds
the threshold is flagged as corrupted.

\begin{table}[htbp]
\centering
\caption{Architecture and training of the denoising autoencoder.}
\label{tab:ae}
\begin{tabular}{ll}
\toprule
Input window & $256 \times 3$ samples (hop 64) \\
Encoder & 3 conv.\ layers, kernel 7, 16/32/16 channels, stride 1/2/2, ReLU \\
Decoder & 2 transposed conv.\ layers + conv.\ projection to 3 channels \\
Loss / optimizer & Huber loss, AdamW (lr $10^{-3}$, weight decay $10^{-5}$) \\
Training & batch 128, 60 epochs, clean records only \\
Input corruption & Gaussian noise ($\sigma = 0.01$), feature dropout 2\,\% \\
Score / threshold & max.\ window error; 99th percentile of validation scores \\
\bottomrule
\end{tabular}
\end{table}

\subsection{Picking phases}
\label{sec:picking}
To be sure that the records contain P-wave arrivals, we apply the PhaseNet picker~\cite{Zhu19} to them. PhaseNet outputs, for every sample of the record, the probability of a P- and of an S-wave arrival; a pick is declared at a local maximum of this probability exceeding a threshold, for which we used 0.5 instead of the default 0.3 in order to keep only confident picks. A record is discarded if PhaseNet finds no P pick, if it finds more than one pick of the same phase, or if the S pick does not follow the P pick; the S pick serves only for this consistency check. For the 4,568 records with manually picked arrivals the manual picks take precedence over the automatic ones; the remaining records are anchored to the PhaseNet picks. Each retained record is finally cut to the window starting 1~s before the P pick and extending to the end of the triggered record, so that the P arrival is always at the same position and the later phases and the recorded coda are kept in full.

\subsection{Data overview}
Table~\ref{t:summary} shows how many records were removed by the quality checks and how many by the P-wave picking. Relatively many records of natural tectonic events were omitted; this is largely due to the fact that for remote tectonic events only a fragment not containing the P-wave was captured. The strict criteria also remove about half of the mining-induced records, which nevertheless remain by far the largest class. The resulting data set of 39,839 records is used in all experiments below.

\begin{table}
\centering
\caption{Events of the SPF archive by type, the corresponding station records
(five stations per event), the records removed by the quality checks of
Sections~\ref{sec:deterministic}--\ref{sec:learned} and by the P-wave picking
of Section~\ref{sec:picking}, and the records used in this study. The 85
unclassified events were excluded by type.}
\label{t:summary}
\begin{tabular}{lrrrrr}
\toprule
 & & & \multicolumn{2}{c}{Removed} & \\
\cmidrule(lr){4-5}
Event type & Events & Records & Quality checks & P-wave picking & Used \\
\midrule
Tectonic (T) & 2,516 & 12,580 & 4,292 & 6,424 & \textbf{1,864} \\
Mining-induced (M) & 14,144 & 70,720 & 18,331 & 15,915 & \textbf{36,474} \\
Quarry blast (E) & 639 & 3,195 & 831 & 863 & \textbf{1,501} \\
Other & 85 & 425 & \multicolumn{2}{c}{425 (excluded)} & 0 \\
\midrule
All & 17,384 & 86,920 & 23,454 & 23,202 & \textbf{39,839} \\
\bottomrule
\end{tabular}
\end{table}

\begin{figure}[tbp]
    \centering
    \begin{minipage}[t]{0.49\linewidth}\centering
      \includegraphics[width=\linewidth]{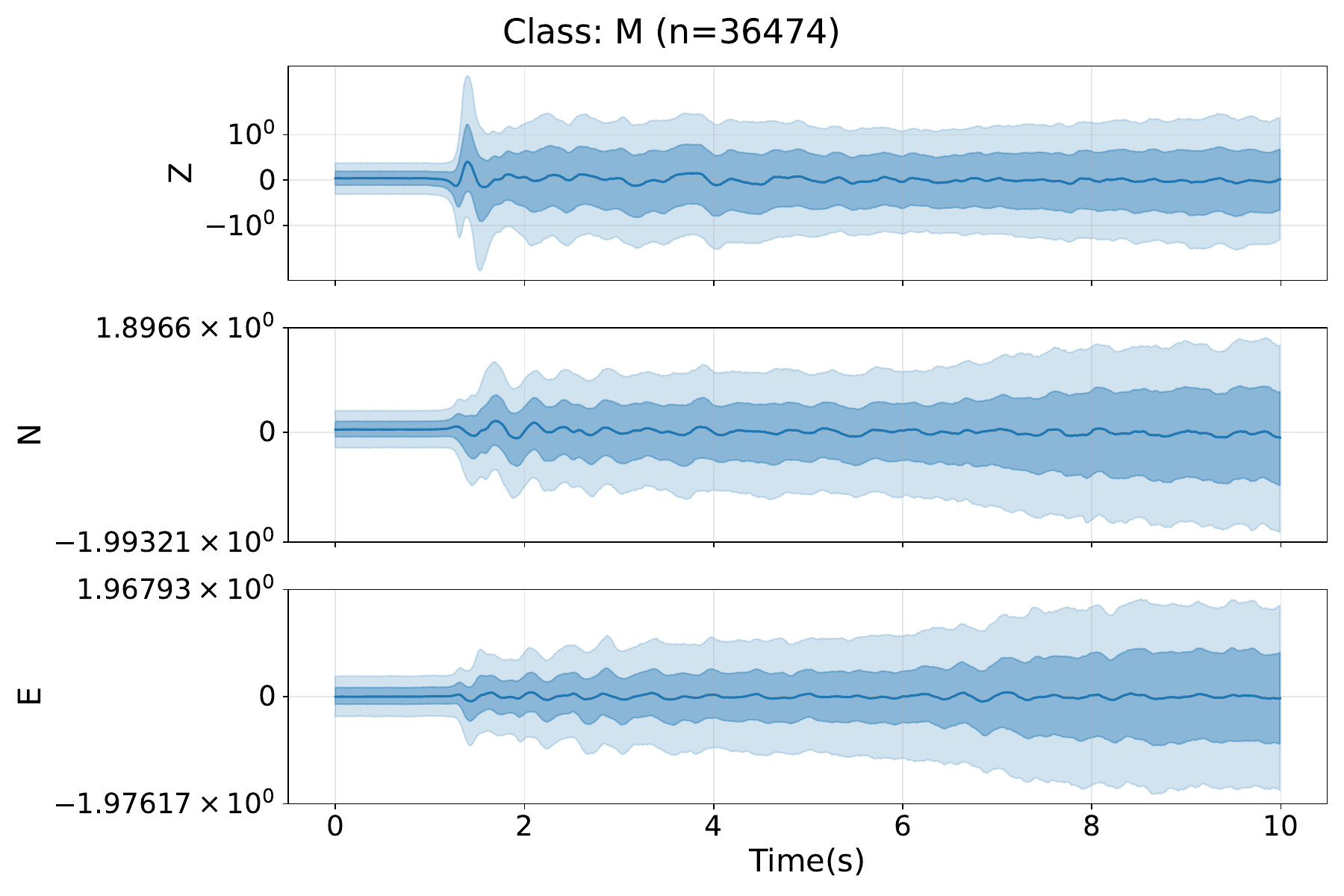}\\[-0.5ex] {\small (a)}
    \end{minipage}\hfill
    \begin{minipage}[t]{0.49\linewidth}\centering
      \includegraphics[width=\linewidth]{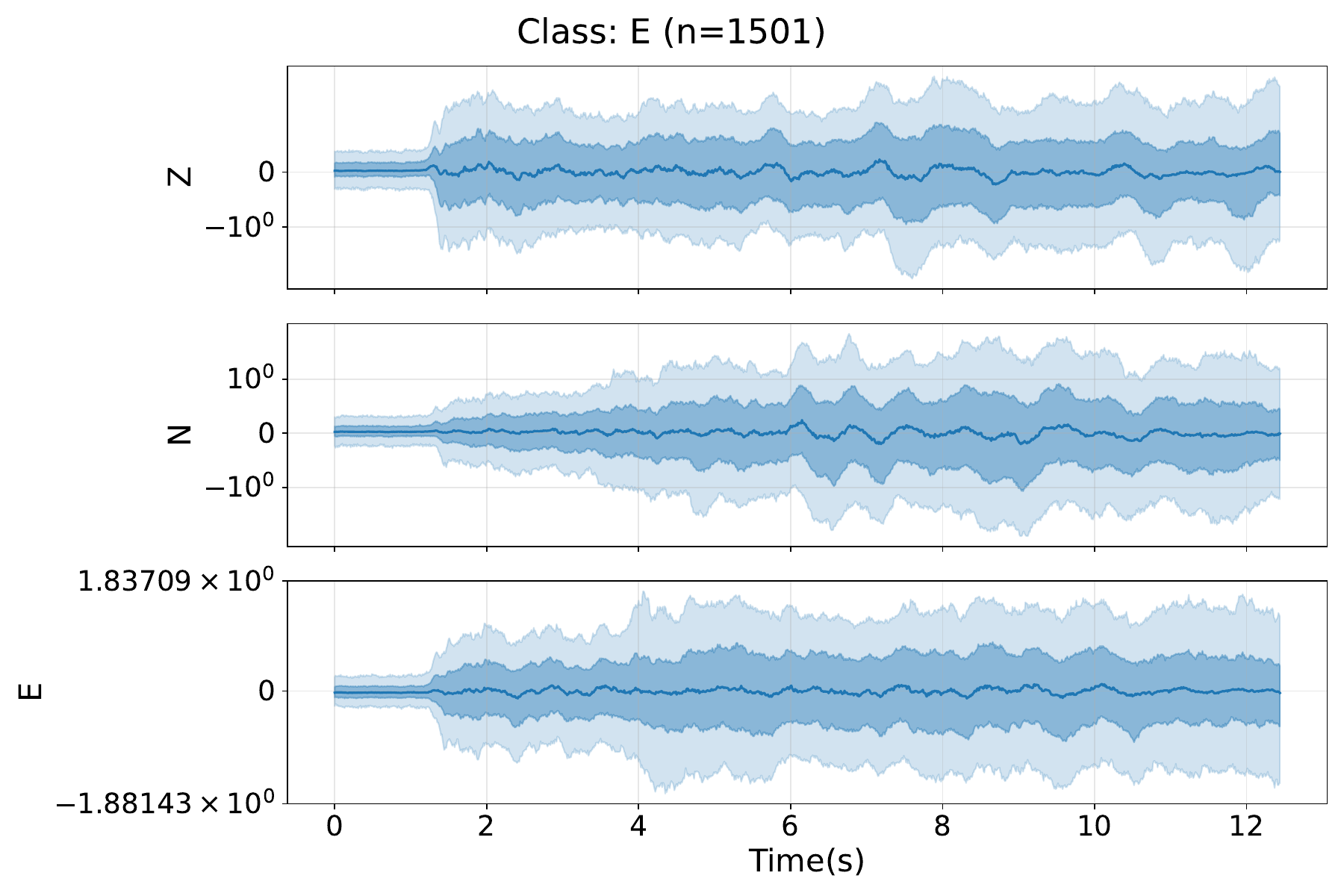}\\[-0.5ex] {\small (b)}
    \end{minipage}\\[1ex]
    \begin{minipage}[t]{0.49\linewidth}\centering
      \includegraphics[width=\linewidth]{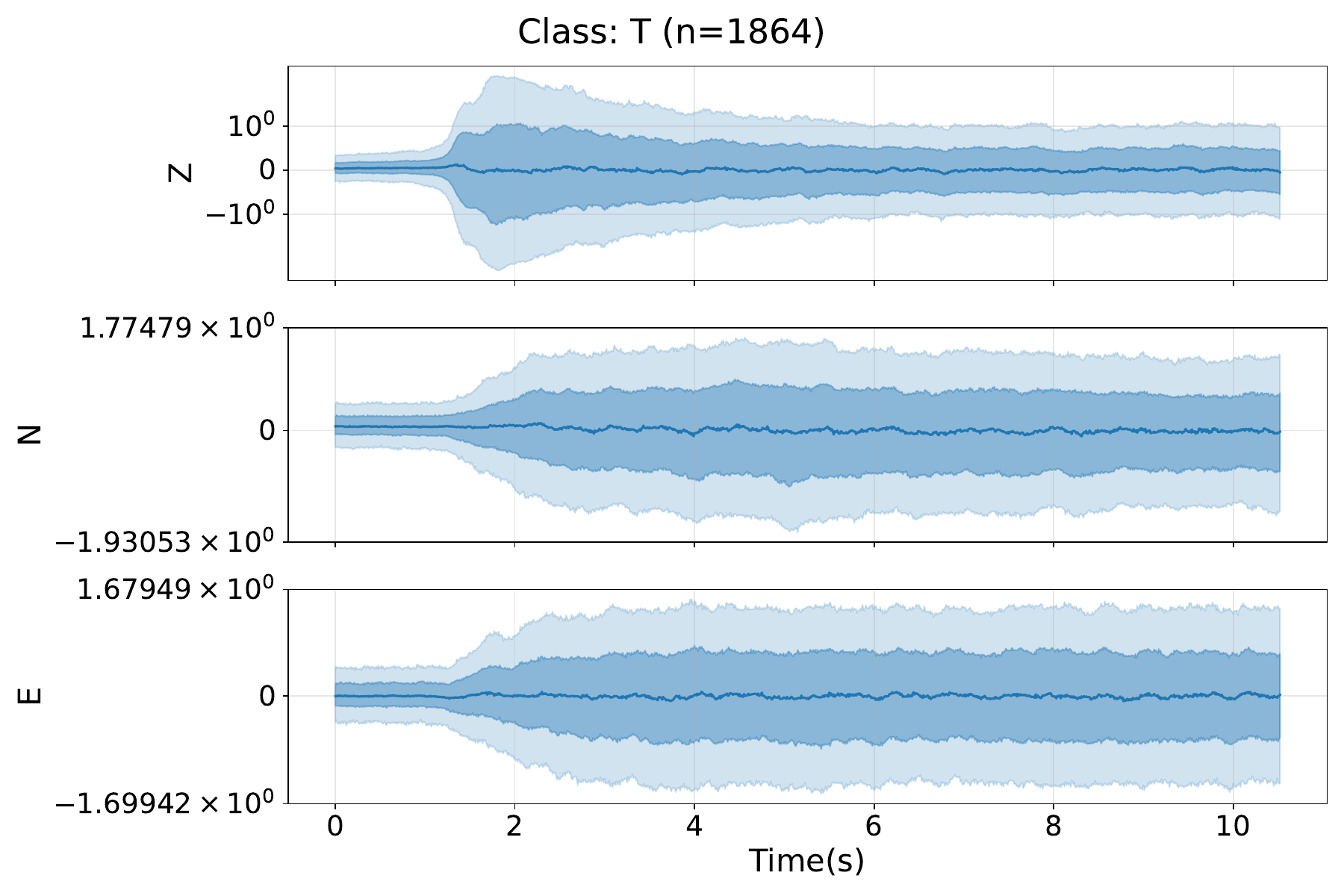}\\[-0.5ex] {\small (c)}
    \end{minipage}
    \caption{Fan charts of the SPF waveforms of (a) mining-induced events,
    (b) quarry blasts and (c) natural tectonic events: median signal (solid
    line) and quantile bands (10--90\,\% and 25--75\,\%) of the z-score
    normalized Z, N and E components.}
    \label{fig:fan_charts}
\end{figure}

Figure~\ref{fig:fan_charts} shows fan charts of the seismic waveforms of natural tectonic (T), mining-induced (M) events, and
quarry blasts (E). Each plot shows the median signal (solid line) and variability across events, represented by quantile bands (10~--~90\% and 25~--~75\%). The differences in the waveform patterns highlight the characteristic features of each type of event.

\section{Neural classification of seismic events}

We classify seismic events into $K$ event types directly from
three-component (Z, N, E) waveform records, without hand-crafted
features. In this work $K = 3$ (blasts, mining-induced and tectonic
events); the formulation below is, however, independent of the number
of classes.

\subsection{Data preprocessing and augmentation}
\label{sec:preprocessing}

Each event is represented by three simultaneously sampled channels
(125\,Hz).\footnote{The sample counts given in this section refer to the
125-Hz SPF archive. The same procedure, with the durations converted to
samples accordingly, is applied in Section~\ref{sec:transfer} to the
100-Hz OKC records and to the SPF records resampled to 100\,Hz.}
The records start 1 s before the P-wave onset and their
length varies between 8 and 64\,s (median 24\,s). To obtain inputs of a
fixed size, every record is cropped to a fixed duration $D$ measured
from its beginning, so that the pre-onset noise and the P arrival are
always retained; records shorter than $D$ are zero-padded at the end.
Each channel is z-score normalized per event (Eq.~(\ref{eq:zscore})) before padding, so that
the padded samples remain exactly zero and do not bias the statistics.
The duration $D$ is treated as a hyperparameter and selected separately
for the two model families (Section~\ref{sec:selection}): $D = 8$\,s
(1,000 samples) for the recurrent models and $D = 24$\,s (3,000 samples)
for the spectrogram models. The former equals the length of the shortest
record, so the recurrent models never see padded input; the latter equals
the median record length, so about half of the spectrogram inputs are
partly zero-padded.
The normalized waveforms are then converted to one of two spectral
representations described in Section~\ref{sec:tf-repr}.

To improve generalization, augmentations are applied on the fly during
training, in the time domain before the spectral transform: random time
shifts, additive noise at a controlled signal-to-noise ratio, amplitude
scaling, and time stretching; spectrogram-based models additionally use
random erasing on the transformed input. The validation and test data
are left unaugmented.

\subsection{Time--frequency representations}
\label{sec:tf-repr}

Seismic event types differ both in their spectral content --- blasts,
for instance, are typically richer in high frequencies than tectonic
events of comparable magnitude --- and in how that content evolves over
the duration of the record. We therefore employ two complementary
Fourier representations, matched to the two families of architectures.

\paragraph{Magnitude spectrum (FFT).}
For the LSTM and the LSTM with self-attention, each channel $c$ of an event is reduced to its
one-sided magnitude spectrum. With an $N$-point discrete Fourier
transform, where $N$ is the smallest power of two not smaller than the
record length ($N = 1,024$ for $D = 8$\,s, i.e.\ a bin width of
0.12\,Hz),
\begin{equation}
  X_c[m] = \left| \sum_{t=0}^{N-1} x_c[t]\,
    e^{-\mathrm{i}\, 2\pi m t / N} \right|,
  \qquad m = 0, \dots, N/2,
  \label{eq:fft}
\end{equation}
yielding $N/2 + 1 = 513$ frequency bins per channel. Discarding the
phase makes the representation insensitive to the exact onset time of
the event within the record, while the magnitudes retain the spectral
shape that discriminates between event types. The three per-channel
spectra are stacked into a $513 \times 3$ array, which these two
models consume as a sequence along the frequency axis --- the recurrent
network thus models dependencies across frequency rather than across
time. Longer records were also evaluated ($D = 16$ and $24$\,s); they
double or quadruple the length of the frequency sequence and
consistently degraded the recurrent models, so $D = 8$\,s is used.

\paragraph{Spectrogram (STFT).}
The convolutional and transformer models instead operate on the
short-time Fourier transform, which preserves the temporal evolution of
the spectrum. Each channel is windowed with a Hann window $w$ of length
$L = 64$ samples (0.51\,s) advanced with hop $H = 32$ (0.26\,s, 50\,\%
overlap), and the
log-compressed magnitude is taken:
\begin{equation}
  S_c[m, \tau] = \log\!\left(1 + \left| \sum_{t=0}^{L-1}
    x_c[t + \tau H]\, w[t]\, e^{-\mathrm{i}\, 2\pi m t / L} \right|\right),
  \qquad m = 0, \dots, L/2,
  \label{eq:stft}
\end{equation}
where $\tau$ indexes the time frame. The logarithm compresses the large
dynamic range between the P and S arrivals and the coda. This produces
a spectrogram with $L/2 + 1 = 33$ frequency bins per channel (bin width
1.95\,Hz up to the Nyquist frequency of 62.5\,Hz); the three
channels are stacked into a three-channel image of shape
$3 \times 33 \times T$, with $T$ the number of time frames ($T = 95$ for
$D = 24$\,s), and processed by the image architectures exactly as an
RGB image would be. The short window trades frequency resolution for
time resolution, which suits the transient, non-stationary character of
seismic events: the arrivals of distinct wave phases and the decay of
the coda remain visible along the time axis. In contrast to the
recurrent models, the spectrogram models benefit from longer records
(validation macro-F1 increases monotonically from $D = 8$ to
$24$\,s), which is consistent with the S-wave arrival and the coda
carrying information that a whole-record magnitude spectrum discards.

\subsection{Model architectures}

Five architectures are compared. Two recurrent variants operate on the
FFT sequence and a third, the LSTM-FCN, on the raw waveform, whereas
EfficientNet and the hybrid CNN--ViT process the STFT spectrogram. This selection allows us to compare a purely sequential model,
sequential models with global attention or with a parallel convolutional branch,
and two image-based models that learn local and global time--frequency
patterns. All models end in a linear layer with $K$ outputs
and are trained with the (unweighted) cross-entropy loss.

Seismic catalogues are typically imbalanced --- in our data one event
type accounts for more than 90\,\% of the events --- and a common remedy
is to weight the loss by inverse class frequencies,
\begin{equation}
  w_k = \frac{N}{K \, n_k},
  \label{eq:class-weights}
\end{equation}
where $N$ is the number of training events and $n_k$ the number of
events of class $k$. We evaluated this weighting, a softened variant
($\sqrt{w_k}$), and random undersampling of the majority class with the
LSTM, and found all of them to \emph{reduce} the macro-averaged F1 score
on the validation split (Section~\ref{sec:selection}): the strongly up-weighted
minority classes absorb a few percent of the majority class, which, given
the imbalance, already outnumber the true minority events, and the
weighted losses also made training noticeably less stable. All reported
models are therefore trained on the natural class distribution; the
imbalance is accounted for in the evaluation metric instead.

\subsubsection{LSTM classifier}

The baseline recurrent model is a stack of three unidirectional long
short-term memory (LSTM) layers~\citep{LSTM97} with a hidden dimension of 128. At frequency
index $t$, an LSTM cell updates its memory $\mathbf{c}_t$ and hidden state
$\mathbf{h}_t$ using input, forget, and output gates. In compact form,
\begin{align}
  \mathbf{i}_t &= \sigma(W_i\mathbf{x}_t + U_i\mathbf{h}_{t-1}+\mathbf{b}_i),
  &\mathbf{f}_t &= \sigma(W_f\mathbf{x}_t + U_f\mathbf{h}_{t-1}+\mathbf{b}_f),\\
  \mathbf{o}_t &= \sigma(W_o\mathbf{x}_t + U_o\mathbf{h}_{t-1}+\mathbf{b}_o),
  &\widetilde{\mathbf{c}}_t &= \tanh(W_c\mathbf{x}_t
    +U_c\mathbf{h}_{t-1}+\mathbf{b}_c),\\
  \mathbf{c}_t &= \mathbf{f}_t\odot\mathbf{c}_{t-1}
    +\mathbf{i}_t\odot\widetilde{\mathbf{c}}_t,
  &\mathbf{h}_t &= \mathbf{o}_t\odot\tanh(\mathbf{c}_t).
\end{align}
Here, $\mathbf{x}_t\in\mathbb{R}^3$ contains the spectral magnitudes of the
three components at frequency index $t$; $\mathbf{i}_t$, $\mathbf{f}_t$
and $\mathbf{o}_t$ are the input, forget and output gates,
$\widetilde{\mathbf{c}}_t$ is the candidate memory, $W_\ast$ and
$U_\ast$ are the learned weight matrices acting on the input and on the
previous hidden state, $\mathbf{b}_\ast$ are the biases, $\sigma$ is the
logistic sigmoid function and $\odot$ denotes the element-wise product;
all gate and state vectors have the hidden dimension 128. Dropout with
probability 0.3 is applied between recurrent layers. The final hidden state of the uppermost layer summarizes the complete
frequency sequence; it is batch-normalized and projected to $K$ class logits
by a fully connected layer. This model provides a simple reference for
measuring the benefit of explicit attention and image-based processing.

\subsubsection{LSTM with multi-head self-attention}

The second model retains the full sequence of LSTM outputs instead of using
only its final state. It uses the same recurrent backbone as the baseline
--- three LSTM layers with hidden dimension 128 and dropout 0.3 --- so that
the comparison isolates the effect of the attention mechanism. The LSTM
output sequence is batch-normalized~\citep{Ioffe2015} and then processed by four-head
self-attention~\citep{Vaswani2017}, in which the same sequence supplies queries, keys, and
values. For one attention head,
\begin{equation}
  \operatorname{Attention}(Q,K,V)
  = \operatorname{softmax}\!\left(\frac{QK^{\mathsf T}}{\sqrt{d_k}}\right)V,
  \label{eq:self-attention}
\end{equation}
where $Q$, $K$ and $V$ are linear projections of the input sequence to
queries, keys and values, $d_k$ is their dimension (the hidden dimension
divided by the number of heads, i.e.\ 32), and the softmax is taken over
the sequence positions; the outputs of the four heads are concatenated
and linearly transformed.
Consequently, every frequency position can directly incorporate information
from every other position. A residual connection adds the normalized LSTM
sequence to the attention output, after which layer normalization is applied.
Mean pooling over all sequence positions produces a fixed-dimensional event
representation, and a linear layer maps it to the class logits. Compared with
the baseline LSTM, this architecture is less dependent on the information
retained at the final sequence position and can model widely separated
spectral features explicitly. We note that this model is considerably more
sensitive to the optimization settings than the baseline: with the learning
rate used for the other networks ($10^{-3}$) it repeatedly collapsed to
predicting the majority class, irrespective of gradient clipping or reduced
dropout, and a ten times smaller learning rate was required for stable
training (Section~\ref{sec:selection}).

\subsubsection{LSTM-FCN}

The third recurrent model, the LSTM fully convolutional network
(LSTM-FCN)~\citep{lstm_fcn_karim}, adds a convolutional branch in parallel
to the recurrent one instead of an attention layer on top of it. Unlike
the two previous models it reads the raw z-score-normalized waveform
($1,000 \times 3$ samples for $D = 8$\,s) rather than its spectrum: on
the magnitude spectrum it was slightly worse, and its small variant
considerably worse (Table~\ref{tab:ablation}d), in agreement with the
observation of the thesis mentioned below. The recurrent branch is a
single LSTM layer with 128 hidden units whose final hidden state, after
dropout with probability 0.8, summarizes the time series. The
convolutional branch reads the same $1,000 \times 3$ sequence with the
three components as input channels and consists of three blocks of a
one-dimensional convolution (128, 256 and 128 filters of width 8, 5 and
3), batch normalization and ReLU, followed by global average pooling over
the time axis, which yields a 128-dimensional descriptor of local
waveform patterns irrespective of their position. The two descriptors are concatenated and projected to
$K$ logits by a linear layer. This architecture was first applied to
the classification of seismic events in the region, in a binary setting,
in the bachelor thesis of one of the authors~\citep{Rieznikov2025BT},
whose implementation used 16 hidden units and 16 filters per block; it
performed very well in our earlier binary studies, quarry blasts versus
tectonic events on the station OKC~\citep{Skotnica2025medgu} and quarry
blasts versus mining-induced events on the stations MORC and
VRAC~\citep{Skotnica2026egu}. Both a variant of this
size and the configuration of the original paper were evaluated here
(Section~\ref{sec:selection}).

\subsubsection{EfficientNet}

EfficientNet-B0~\citep{TanLe2019} is used as a convolutional spectrogram classifier. Its
ImageNet-pretrained feature extractor consists of mobile inverted bottleneck
convolutions with squeeze-and-excitation blocks. The architecture jointly
scales network depth, width, and input resolution, providing a favorable
accuracy--cost trade-off. Because the three seismic components form a
three-channel input, the pretrained input layer can be used without changing
its channel count. The original ImageNet classification head is removed;
features are aggregated by global average pooling and passed through dropout
with probability 0.2 and a new $K$-output linear layer. The complete network
is fine-tuned, allowing early filters to adapt from natural-image structure to
local time--frequency patterns such as arrivals, narrow frequency bands, and
coda decay.

\subsubsection{Hybrid convolutional vision transformer}

The hybrid CNN--ViT combines a convolutional front end with the first four
blocks of an ImageNet-pretrained ViT-Tiny transformer~\citep{Dosovitskiy2021}. The convolutional stem
contains three stride-two convolutions: a $7\times7$ layer with 64 output
channels, a $3\times3$ layer with 128 channels, and a $3\times3$ layer whose
width equals the transformer embedding dimension. Batch normalization follows
each convolution, and the first two layers additionally use ReLU activation.
The stem reduces both spectrogram dimensions by approximately a factor of
eight while learning local features more gradually than a single non-
overlapping patch projection.

The resulting feature map is flattened into a sequence of patch tokens. A
learnable classification token is prepended, and positional embeddings from
the pretrained transformer are bicubically interpolated to the spectrogram's
patch grid. After dropout with probability 0.2, the token sequence passes
through four transformer blocks, each combining multi-head self-attention and
a feed-forward network with residual connections. The normalized
classification token is finally projected to $K$ logits. The convolutional
stem supplies locality and translation bias, while self-attention captures
relations between distant regions of the time--frequency plane.

\subsection{Training and model selection}
\label{sec:selection}

The 39,839 records of the cleaned archive (Section~\ref{sec:cleaning}),
each of which is classified independently and is referred to as an
event in the following, are split into training, validation, and test
sets in the proportion 56\,:\,14\,:\,30\,\%, stratified by class, i.e.\
the test set holds 30\,\% of the records and the validation set 20\,\% of
the remaining 70\,\%. The split is made at the level of records, so
that the records of one seismic event from different stations may fall
into different subsets; after cleaning, many events are represented by
fewer than five records. All networks are trained with the AdamW optimizer~\citep{Loshchilov2019} and
an exponentially decaying learning rate (multiplied by $\gamma$ after
every epoch) for a fixed budget of 30 epochs, in mixed precision, with
mini-batches of 32 events for the recurrent and 64 events for the
spectrogram models. The initial learning rate is $10^{-3}$ with
$\gamma = 0.85$ for the LSTM, LSTM-FCN, EfficientNet, and CNN--ViT models,
and $10^{-4}$ with $\gamma = 0.9$ for the LSTM with self-attention. The
pretrained spectrogram models converge within roughly 20 epochs; the
recurrent models are still improving slowly at the end of the budget.

All design choices discussed above --- the record duration $D$, the
logarithmic compression of the spectrum or spectrogram, the class
weighting, the learning rate of the LSTM with self-attention, and the
input representation and size of the LSTM-FCN --- were selected on
the validation split by the macro-averaged F1 score, averaged over three
training runs with different random seeds. The test split was not used
for any of these decisions. For each architecture, the selected
configuration was then retrained on the union of the training and
validation sets (70\,\% of the events) with the same epoch budget and
evaluated once on the test split. We report the mean and standard
deviation over three seeds of these final models.

\subsection{Evaluation}

Models are evaluated on the held-out test split using the
macro-averaged F1 score, i.e.\ the unweighted mean of the per-class F1
scores. Macro averaging treats all classes equally regardless of their
numbers, which is appropriate for imbalanced seismic catalogues: a
classifier that labels every event with the majority class attains an
accuracy above 90\,\% on our data but a macro-F1 of only 0.32. Per-class
F1 scores and confusion matrices are additionally inspected to identify
which event types the models confuse.

\section{Results on the SPF archive}
\label{sec:results}

\begin{table}[tbp]
  \centering
  \caption{Test-set performance of the final models (trained on 70\,\% of the events, evaluated on the held-out 30\,\%). Mean $\pm$ standard deviation over three training runs with different random seeds. $D$ is the record duration, Acc.\ the overall accuracy, and the last columns give the per-class F1 scores.}
  \label{tab:final-results}
  \begin{tabular}{llrrrccc}
    \toprule
    Model & Input & $D$ [s] & Params & Macro-F1 & F1$_\mathrm{E}$ & F1$_\mathrm{M}$ & F1$_\mathrm{T}$ \\
    \midrule
    EfficientNet-B0 & STFT & 24 & 4.05\,M & 0.961 $\pm$ 0.002 & 0.946 & 0.996 & 0.941 \\
    Hybrid CNN--ViT & STFT & 24 & 2.10\,M & 0.947 $\pm$ 0.002 & 0.922 & 0.994 & 0.923 \\
    LSTM & FFT & 8 & 333\,k & 0.887 $\pm$ 0.005 & 0.822 & 0.988 & 0.851 \\
    LSTM-FCN & waveform & 8 & 337\,k & 0.885 $\pm$ 0.005 & 0.809 & 0.989 & 0.858 \\
    LSTM + self-attention & FFT & 8 & 399\,k & 0.833 $\pm$ 0.004 & 0.703 & 0.984 & 0.811 \\
    \bottomrule
  \end{tabular}
\end{table}

\begin{table}[tbp]
  \centering
  \caption{Per-class precision and recall of the final models on the test split (mean over three seeds) and overall accuracy. For comparison, the majority-class baseline (every event labelled M) has an accuracy of 0.916 and a macro-F1 of 0.319.}
  \label{tab:prec-rec}
  \begin{tabular}{lcccccccc}
    \toprule
    & \multicolumn{2}{c}{E} & \multicolumn{2}{c}{M} & \multicolumn{2}{c}{T} & \\
    \cmidrule(lr){2-3} \cmidrule(lr){4-5} \cmidrule(lr){6-7}
    Model & Prec. & Rec. & Prec. & Rec. & Prec. & Rec. & Acc. \\
    \midrule
    EfficientNet-B0 & 0.956 & 0.936 & 0.995 & 0.996 & 0.948 & 0.933 & 0.991 \\
    Hybrid CNN--ViT & 0.935 & 0.910 & 0.993 & 0.996 & 0.943 & 0.904 & 0.988 \\
    LSTM & 0.858 & 0.789 & 0.984 & 0.993 & 0.906 & 0.803 & 0.976 \\
    LSTM-FCN & 0.864 & 0.760 & 0.983 & 0.996 & 0.935 & 0.792 & 0.977 \\
    LSTM + self-attention & 0.852 & 0.599 & 0.972 & 0.997 & 0.965 & 0.699 & 0.968 \\
    \bottomrule
  \end{tabular}
\end{table}

\begin{table}[tbp]
  \centering
  \caption{Confusion matrices of the final models on the test split (450~E, 10,943~M, 559~T events), averaged over three seeds and rounded to whole events. Rows are true classes, columns predicted classes.}
  \label{tab:confusion}
  \begin{tabular}{llrrr}
    \toprule
    Model & True & \multicolumn{1}{c}{$\hat{\mathrm{E}}$} & \multicolumn{1}{c}{$\hat{\mathrm{M}}$} & \multicolumn{1}{c}{$\hat{\mathrm{T}}$} \\
    \midrule
    \multirow{3}{*}{EfficientNet-B0} & E & \textbf{421} & 27 & 2 \\
     & M & 13 & \textbf{10903} & 27 \\
     & T & 6 & 31 & \textbf{522} \\
    \midrule
    \multirow{3}{*}{Hybrid CNN--ViT} & E & \textbf{410} & 36 & 4 \\
     & M & 20 & \textbf{10897} & 26 \\
     & T & 9 & 45 & \textbf{505} \\
    \midrule
    \multirow{3}{*}{LSTM} & E & \textbf{355} & 86 & 9 \\
     & M & 41 & \textbf{10865} & 38 \\
     & T & 18 & 92 & \textbf{449} \\
    \midrule
    \multirow{3}{*}{LSTM-FCN} & E & \textbf{342} & 95 & 13 \\
     & M & 30 & \textbf{10895} & 18 \\
     & T & 24 & 92 & \textbf{443} \\
    \midrule
    \multirow{3}{*}{LSTM + self-attention} & E & \textbf{270} & 175 & 6 \\
     & M & 22 & \textbf{10912} & 8 \\
     & T & 25 & 143 & \textbf{391} \\
    \bottomrule
  \end{tabular}
\end{table}

\begin{table}[tbp]
  \centering
  \caption{Model selection on the validation split: macro-F1 (mean $\pm$ std over three seeds) and per-class F1 for the design choices discussed in the text. Unless stated otherwise, models are trained without class weights, spectrogram models with log compression, and recurrent models on the linear magnitude spectrum.}
  \label{tab:ablation}
  \setlength{\tabcolsep}{4pt}
  \begin{tabular}{lcccc}
    \toprule
    Configuration & Macro-F1 & F1$_\mathrm{E}$ & F1$_\mathrm{M}$ & F1$_\mathrm{T}$ \\
    \midrule \multicolumn{5}{l}{\emph{(a) Record duration $D$}} \\
    EfficientNet-B0, $D = 8$\,s & 0.927 $\pm$ 0.006 & 0.926 & 0.992 & 0.864 \\
    EfficientNet-B0, $D = 16$\,s & 0.947 $\pm$ 0.002 & 0.945 & 0.994 & 0.903 \\
    EfficientNet-B0, $D = 24$\,s & 0.951 $\pm$ 0.000 & 0.940 & 0.995 & 0.919 \\
    Hybrid CNN--ViT, $D = 8$\,s & 0.923 $\pm$ 0.006 & 0.920 & 0.991 & 0.857 \\
    Hybrid CNN--ViT, $D = 16$\,s & 0.947 $\pm$ 0.002 & 0.936 & 0.994 & 0.912 \\
    Hybrid CNN--ViT, $D = 24$\,s & 0.949 $\pm$ 0.001 & 0.935 & 0.995 & 0.918 \\
    LSTM, $D = 8$\,s & 0.880 $\pm$ 0.008 & 0.836 & 0.987 & 0.817 \\
    LSTM, $D = 16$\,s & 0.858 $\pm$ 0.003 & 0.774 & 0.985 & 0.814 \\
    LSTM, $D = 24$\,s & 0.843 $\pm$ 0.006 & 0.746 & 0.983 & 0.801 \\
    \midrule \multicolumn{5}{l}{\emph{(b) Logarithmic compression of the spectrum or spectrogram ($D = 16$\,s)}} \\
    EfficientNet-B0, linear & 0.948 $\pm$ 0.003 & 0.944 & 0.995 & 0.907 \\
    EfficientNet-B0, log & 0.947 $\pm$ 0.002 & 0.945 & 0.994 & 0.903 \\
    Hybrid CNN--ViT, linear & 0.942 $\pm$ 0.003 & 0.929 & 0.994 & 0.904 \\
    Hybrid CNN--ViT, log & 0.947 $\pm$ 0.002 & 0.936 & 0.994 & 0.912 \\
    LSTM, linear & 0.858 $\pm$ 0.003 & 0.774 & 0.985 & 0.814 \\
    LSTM, log & 0.826 $\pm$ 0.006 & 0.707 & 0.983 & 0.787 \\
    \midrule \multicolumn{5}{l}{\emph{(c) Class imbalance (LSTM, $D = 16$\,s)}} \\
    No weighting & 0.858 $\pm$ 0.003 & 0.774 & 0.985 & 0.814 \\
    Weights $\sqrt{w_k}$ & 0.848 $\pm$ 0.010 & 0.775 & 0.982 & 0.788 \\
    Weights $w_k$ (inverse frequency) & 0.757 $\pm$ 0.014 & 0.626 & 0.961 & 0.683 \\
    Undersampling M to 2,000 events & 0.752 $\pm$ 0.001 & 0.618 & 0.962 & 0.676 \\
    \midrule \multicolumn{5}{l}{\emph{(d) Extended recurrent models}} \\
    LSTM + attn., 8 layers $\times$ 64, lr $10^{-3}$, $D = 16$\,s & 0.319 $\pm$ 0.000 & 0.000 & 0.956 & 0.000 \\
    LSTM + attn., 3 layers $\times$ 128, lr $3\cdot10^{-4}$, $D = 16$\,s & 0.316 $\pm$ 0.170 & 0.199 & 0.613 & 0.137 \\
    LSTM + attn., 3 layers $\times$ 128, lr $10^{-4}$, $D = 16$\,s & 0.792 $\pm$ 0.013 & 0.640 & 0.981 & 0.756 \\
    LSTM + attn., 3 layers $\times$ 128, lr $3\cdot10^{-4}$, $D = 8$\,s & 0.760 $\pm$ 0.103 & 0.533 & 0.979 & 0.768 \\
    LSTM + attn., 3 layers $\times$ 128, lr $10^{-4}$, $D = 8$\,s & 0.813 $\pm$ 0.016 & 0.685 & 0.982 & 0.773 \\
    LSTM-FCN 128 / 128-256-128, spectrum, lr $10^{-3}$, $D = 8$\,s & 0.868 $\pm$ 0.005 & 0.825 & 0.986 & 0.794 \\
    LSTM-FCN 128 / 128-256-128, spectrum, lr $10^{-4}$, $D = 8$\,s & 0.859 $\pm$ 0.002 & 0.807 & 0.985 & 0.785 \\
    LSTM-FCN 16 / 16-16-16, spectrum, lr $10^{-3}$, $D = 8$\,s & 0.720 $\pm$ 0.013 & 0.435 & 0.976 & 0.750 \\
    LSTM-FCN 16 / 16-16-16, spectrum, lr $10^{-4}$, $D = 8$\,s & 0.414 $\pm$ 0.025 & 0.000 & 0.960 & 0.281 \\
    LSTM-FCN 128 / 128-256-128, raw waveform, lr $10^{-3}$, $D = 8$\,s & 0.871 $\pm$ 0.006 & 0.803 & 0.987 & 0.822 \\
    LSTM-FCN 128 / 128-256-128, raw waveform, lr $10^{-3}$, $D = 24$\,s & 0.499 $\pm$ 0.336 & 0.360 & 0.651 & 0.486 \\
    LSTM-FCN 16 / 16-16-16, raw waveform, lr $10^{-3}$, $D = 8$\,s & 0.840 $\pm$ 0.005 & 0.750 & 0.985 & 0.784 \\
    \bottomrule
  \end{tabular}
\end{table}

\begin{table}[tbp]
  \centering
  \caption{Computational cost of the final models: number of trainable parameters, wall-clock training time for 30 epochs, and inference throughput on the test split (11,952 events, including the on-the-fly spectral transform and data loading). Times are means over three runs, each executed alone on one GPU (NVIDIA RTX 5090) with four data-loading workers.}
  \label{tab:cost}
  \begin{tabular}{lrrr}
    \toprule
    Model & Parameters & Training [min] & Inference [events/s] \\
    \midrule
    EfficientNet-B0 & 4,053,456 & 2.6 & 17,158 \\
    Hybrid CNN--ViT & 2,098,182 & 1.4 & 17,758 \\
    LSTM & 333,188 & 3.7 & 28,981 \\
    LSTM-FCN & 336,646 & 3.2 & 24,190 \\
    LSTM + self-attention & 399,492 & 5.0 & 16,835 \\
    \bottomrule
  \end{tabular}
\end{table}

\subsection{Comparison of architectures}

Table~\ref{tab:final-results} summarizes the test-set performance of the
five architectures after retraining with the configurations selected on
the validation split (Section~\ref{sec:selection}). The two
spectrogram-based models clearly outperform the recurrent models, which
operate on the magnitude spectrum or on the raw waveform. EfficientNet-B0 reaches a macro-F1 of
$0.961 \pm 0.002$ and the hybrid CNN--ViT $0.947 \pm 0.002$, whereas the
best recurrent model, the plain LSTM, attains $0.887 \pm 0.005$. The
standard deviations over three training runs are small compared with
the differences between architectures, so the ranking is not an
artefact of a particular random initialization. The majority-class
baseline, which labels every event as M, has an accuracy of 0.916 but a
macro-F1 of only 0.319; all five networks are far above it, and the
differences between them are almost entirely due to the two minority
classes (Table~\ref{tab:prec-rec}).

Within the recurrent family, neither of the two extensions improves on
the baseline. The LSTM-FCN, which reads the raw waveform, matches the
plain LSTM ($0.885 \pm 0.005$ against $0.887 \pm 0.005$) with a slightly
different balance between precision and recall on the minority classes
(precision 0.86 and recall 0.76 on E, 0.94 and 0.79 on T). The LSTM with self-attention
($0.833 \pm 0.004$) is markedly more conservative: its precision on the
minority classes is higher (0.85 and 0.97), but its recall is much lower
(0.60 and 0.70), i.e.\ it misses a large fraction of the blasts and
tectonic events and labels them as mining-induced. Neither a parallel
convolutional branch on the waveform nor an attention layer on the
spectrum thus adds to what the recurrent layers extract.

Retraining on the union of the training and validation sets, i.e.\ on
25\,\% more events than during model selection, improved the test
macro-F1 of EfficientNet from $0.953$ (selection runs trained on the
training split alone) to $0.961$, while the hybrid CNN--ViT ($0.947$)
and the LSTM ($0.883 \rightarrow 0.887$) remained essentially unchanged.
The largest model therefore still benefits from additional data,
suggesting that its performance is limited by the number of minority
events rather than by its capacity.

\subsection{Error structure}

The confusion matrices in Table~\ref{tab:confusion} show that the
remaining errors of all models have the same structure: blasts and
tectonic events are confused with the dominant class M, but hardly
ever with each other. For EfficientNet, about 27 of the 450 test blasts
and 31 of the 559 tectonic events are labelled as mining-induced,
whereas only about two blasts are labelled tectonic and six tectonic
events are labelled as blasts. Conversely, of the 10,943 mining-induced
events only 13 are assigned to E and 27 to T, so the false-alarm rate
on the majority class is below 0.4\,\%. The direct confusion between the
two minority classes is thus negligible; the practically relevant
problem is to separate each of them from the mining-induced events.

The recurrent models exhibit the same pattern with larger numbers. The
LSTM misses about 86 blasts and 92 tectonic events per run, the
LSTM-FCN 95 and 92, and the attention model 175 and 143, almost all
of them absorbed by class M. The attention model is, on the other hand,
the most precise of all models on the tectonic class (0.97,
Table~\ref{tab:prec-rec}), at the price of the lowest recall.
Since we have no reason to regard one kind of confusion as more serious
than another, we judge the models by the macro-F1 score, which weights
precision and recall equally, and by this criterion the spectrogram
models are clearly better.

\subsection{Binary tasks}

For comparison with the binary problems of earlier studies, EfficientNet
and the LSTM were also trained, with their selected configurations, on
each pair of classes alone. EfficientNet separates blasts from
mining-induced events with a macro-F1 of $0.970 \pm 0.002$ (area under
the ROC curve, ROC-AUC, $0.997$), tectonic from mining-induced events
with $0.971 \pm 0.001$ (ROC-AUC $0.995$) and blasts from tectonic
events with $0.971 \pm 0.002$ (ROC-AUC $0.995$); the LSTM reaches
$0.918$, $0.928$ and $0.924$. The ternary EfficientNet, evaluated on
the same pairs of test classes with predictions of the third class
counted as errors, gives $0.961$ for blasts versus mining-induced and
$0.964$ for tectonic versus mining-induced events, so a dedicated
binary classifier gains about $0.01$ on the pairs that involve the
dominant class. For blasts versus tectonic events the comparison is not
meaningful, since nearly all errors of the ternary model on these two
classes are confusions with the mining-induced class, which the binary
task does not contain.

\subsection{Model selection and ablations}

Table~\ref{tab:ablation} reports the validation macro-F1 for the design
choices discussed in the methods section.

\paragraph{Record duration.}
The two model families respond in opposite ways to the duration $D$ of
the analysed record (Table~\ref{tab:ablation}a). For the spectrogram
models, a longer record is consistently better: EfficientNet improves
from $0.927$ at $D = 8$\,s to $0.947$ at 16\,s and $0.951$ at 24\,s,
and the hybrid CNN--ViT from $0.923$ to $0.949$. The gain from 8 to
16\,s is large and concerns above all the tectonic class, whose
S-wave arrival and coda extend beyond the first eight seconds; the gain
from 16 to 24\,s is small, which is expected since half of the records
are shorter than 24\,s and are zero-padded. The LSTM, in contrast,
degrades monotonically from $0.880$ at 8\,s to $0.858$ at 16\,s and
$0.843$ at 24\,s. Because the FFT discards the temporal structure, a
longer record does not add information about the sequence of wave
phases; it only doubles or quadruples the length of the frequency
sequence the recurrent network has to summarize, and the finer
frequency resolution does not compensate for this.

\paragraph{Logarithmic compression.}
Compressing the spectrum with $\log(1 + |\cdot|)$ has little effect on
the spectrogram models (Table~\ref{tab:ablation}b): EfficientNet is
unchanged within one standard deviation and the hybrid CNN--ViT gains
$0.005$. For the LSTM the compression is harmful ($0.826$ versus
$0.858$), presumably because the per-event z-score normalization already
bounds the dynamic range of the input and the logarithm flattens the
spectral peaks that the recurrent model relies on. The log-compressed
spectrogram and the linear magnitude spectrum were therefore used in
the final models.

\paragraph{Class imbalance.}
None of the standard remedies for class imbalance improved the
macro-F1 on the validation split (Table~\ref{tab:ablation}c). Weighting
the cross-entropy loss by the square root of the inverse class
frequencies lowered the LSTM's macro-F1 from $0.858 \pm 0.003$ to
$0.848 \pm 0.010$; full inverse-frequency weighting
(Eq.~\ref{eq:class-weights}, $w_\mathrm{E} = 8.8$, $w_\mathrm{M} = 0.36$,
$w_\mathrm{T} = 7.1$) reduced it to $0.757 \pm 0.014$, and undersampling
the majority class to 2,000 training events to $0.752 \pm 0.001$. The
per-class scores show why. The remedies do what they are designed to do
--- the recall of the minority classes rises from 0.71 to 0.83 (E) and
from 0.73 to 0.86 (T) with inverse weights --- but the decision boundary
moves into the majority class: the number of mining-induced events
labelled as blasts or tectonic events grows from about 30 to about 330
of the 5,107 validation events of class M. Although this is only 6\,\% of
the majority class, it exceeds the number of true minority events in the
validation split (210 E and 261 T), so the minority-class precision
halves (from 0.86 to 0.50 for E and from 0.91 to 0.57 for T) and the
macro-F1 drops by 0.10. The square-root weights produce the same effect
in attenuated form. Training on the natural class distribution and
relying on a class-balanced evaluation metric was therefore the better
choice, and all architectures were trained in this way. Weighting would only be justified if one
kind of confusion were known to be considerably more serious than the
other; at present we have no basis for such an asymmetry and treat all
misclassifications alike.

\paragraph{Extended recurrent models.}
The LSTM with self-attention in its original configuration (eight
layers with 64 hidden units, learning rate $10^{-3}$) failed to learn in
all three runs and predicted the majority class throughout
(Table~\ref{tab:ablation}d, macro-F1 $0.319$). Reducing the depth to the
three-layer, 128-unit backbone of the baseline LSTM was not sufficient:
at a learning rate of $3 \cdot 10^{-4}$ the training still collapsed in
some of the runs (hence the large standard deviations of $0.10$--$0.17$),
and only at $10^{-4}$ did the model train reliably, reaching $0.813$ at
$D = 8$\,s. Additional short diagnostic runs with the three-layer
backbone at $10^{-3}$ (not tabulated) showed that gradient clipping and
a lower dropout rate did not prevent the collapse at the higher learning
rates, so the instability is a property of the attention layer on top of
the recurrent output rather than of exploding gradients alone. The
LSTM-FCN, in contrast, trains without difficulty at the learning rate of
the other models. On the magnitude spectrum it reaches $0.868$ with the
configuration of the original paper (128 hidden units, 128/256/128
filters) and only $0.72$ with a variant of the size used in the thesis
(16 units and 16 filters per block); on the raw waveform both variants improve, to $0.871$ and
$0.84$, which agrees with the observation of the thesis that the
architecture works best on the time series itself, and the waveform
input was therefore selected. A longer raw window of 24\,s (3,000 time
steps) was not learnable for the recurrent branch ($0.50 \pm 0.34$).
Even with their best settings, however, both extended models remain at
or below the plain LSTM ($0.880$), confirming the test-set ranking of
Table~\ref{tab:final-results}.

\subsection{Computational cost}

Table~\ref{tab:cost} lists the number of parameters, the training time
and the inference throughput of the final models, measured with each run
executed alone on one GPU. The spectrogram models are larger --- 4.1
million parameters for EfficientNet-B0 and 2.1 million for the hybrid
CNN--ViT, compared with 0.33--0.40 million for the recurrent models ---
but they are faster to train: the hybrid CNN--ViT needs 1.4 and
EfficientNet 2.6 minutes for 30 epochs, against 3.7 minutes for the LSTM
and 5 minutes for the attention model, because the convolutional
networks process the $33 \times 95$ spectrogram in parallel, whereas the
recurrent networks have to traverse a sequence of 513 frequency bins step
by step; the LSTM-FCN needs 3.2 minutes, its single recurrent layer
traversing the 1,000 samples of the waveform. At inference, all models classify the 11,952 test events in
under a second, i.e.\ at more than $10^4$ events per second including the
spectral transform, so that the computational cost of the classifier is
irrelevant for an operational system, in which the event rate is orders
of magnitude lower. The hybrid CNN--ViT thus offers a favourable
compromise, with a macro-F1 within $0.015$ of EfficientNet at half the
parameters and half the training time.

\section{Transfer to the present-day station OKC}
\label{sec:transfer}

\begin{table}[tbp]
  \centering
  \caption{Ternary classification on the OKC test events (699~E, 5,379~M, 46~T) with EfficientNet-B0 trained in four ways; mean $\pm$ std over three seeds. ``SPF test'' is the same model evaluated on the SPF part of the test split (450~E, 10,942~M, 559~T). The record length $D$ is the one selected on the validation split for each strategy; fine-tuning starts from the SPF-only model and continues on the OKC training events for 15 epochs at a learning rate of $10^{-4}$.}
  \label{tab:transfer-strategies}
  \begin{tabular}{llcccccc}
    \toprule
    Training data & $D$ [s] & \multicolumn{4}{c}{OKC test} & SPF test \\
    \cmidrule(lr){3-6} \cmidrule(lr){7-7}
    & & Macro-F1 & F1$_\mathrm{E}$ & F1$_\mathrm{M}$ & F1$_\mathrm{T}$ & Macro-F1 \\
    \midrule
    OKC only & 16 & 0.915 $\pm$ 0.005 & 0.975 & 0.996 & 0.773 & 0.288 $\pm$ 0.014 \\
    SPF + OKC (pooled) & 20 & 0.918 $\pm$ 0.007 & 0.975 & 0.996 & 0.783 & 0.961 $\pm$ 0.001 \\
    SPF, fine-tuned on OKC & 20 & 0.864 $\pm$ 0.007 & 0.958 & 0.994 & 0.639 & 0.393 $\pm$ 0.021 \\
    SPF only (no OKC data) & 20 & 0.387 $\pm$ 0.006 & 0.686 & 0.453 & 0.021 & 0.958 $\pm$ 0.001 \\
    \bottomrule
  \end{tabular}
\end{table}

\begin{table}[tbp]
  \centering
  \caption{Model selection for the OKC station: macro-F1 of EfficientNet-B0 on the OKC part of the validation split as a function of the record length $D$ and of the training data (selection runs, trained on the training split only; mean $\pm$ std over three seeds). The last column gives the number of the 21 OKC tectonic validation events that are recognised (rounded mean over the seeds).}
  \label{tab:transfer-length}
  \begin{tabular}{llccc}
    \toprule
    Training data & $D$ [s] & Macro-F1 (OKC) & F1$_\mathrm{T}$ (OKC) & T recognised \\
    \midrule
    OKC only & 8 & 0.840 $\pm$ 0.015 & 0.601 & 11 / 21 \\
    OKC only & 16 & 0.898 $\pm$ 0.008 & 0.748 & 15 / 21 \\
    OKC only & 20 & 0.892 $\pm$ 0.004 & 0.736 & 15 / 21 \\
    \midrule
    SPF + OKC & 8 & 0.873 $\pm$ 0.017 & 0.691 & 13 / 21 \\
    SPF + OKC & 16 & 0.891 $\pm$ 0.005 & 0.724 & 15 / 21 \\
    SPF + OKC & 20 & 0.914 $\pm$ 0.009 & 0.789 & 16 / 21 \\
    \bottomrule
  \end{tabular}
\end{table}

\begin{table}[tbp]
  \centering
  \caption{Confusion matrices on the OKC test events (rows: true class, columns: predicted; averaged over three seeds and rounded) for the pooled model and for the SPF-only model applied to OKC without any OKC training data.}
  \label{tab:transfer-confusion}
  \begin{tabular}{llrrr}
    \toprule
    Model & True & \multicolumn{1}{c}{$\hat{\mathrm{E}}$} & \multicolumn{1}{c}{$\hat{\mathrm{M}}$} & \multicolumn{1}{c}{$\hat{\mathrm{T}}$} \\
    \midrule
    \multirow{3}{*}{SPF + OKC (pooled)} & E & \textbf{674} & 23 & 2 \\
     & M & 7 & \textbf{5370} & 2 \\
     & T & 2 & 12 & \textbf{32} \\
    \midrule
    \multirow{3}{*}{SPF only} & E & \textbf{473} & 43 & 182 \\
     & M & 216 & \textbf{1595} & 3568 \\
     & T & 2 & 4 & \textbf{39} \\
    \bottomrule
  \end{tabular}
\end{table}

The results of the previous section were obtained on records of a
network that stopped operating in 2002. The practical question is
whether they carry over to the station that monitors the region today.
We therefore repeat the ternary experiment with the OKC data set
(Section~\ref{sec:okc-data}) in three settings: training on the OKC
data alone, training on the SPF archive alone and applying the model to
OKC without any adaptation, and training on the union of both data
sets. In addition we test the common recipe of pre-training on the
large archive and fine-tuning on the smaller modern data set. Since the
spectrogram-based EfficientNet was the most accurate model on the SPF
data, the comparison is restricted to this architecture.

\subsection{Setup}

The two data sets differ in sampling rate (125\,Hz for SPF, 100\,Hz for
OKC), so the SPF records were resampled to 100\,Hz with a polyphase
filter including anti-aliasing. The amplitude units differ as well (the
OKC records are raw digital counts); the per-event z-score normalization
of Section~\ref{sec:preprocessing} removes this difference, but not the
difference in the instrument responses. Records are cropped to $D = 8$,
16 or 20\,s from one second before the P onset as before; all OKC
records and more than 80\,\% of the SPF records are at least 20\,s long,
so zero-padding plays a minor role.

The events of both sources were split jointly into training, validation
and test sets (56\,:\,14\,:\,30\,\%), stratified by source and class, so
that the OKC part of the test split (699~E, 5,379~M and 46~T events)
is the same for every model and the SPF part coincides, up to the
resampling and a difference of one event due to the joint
stratification, with the test split of Section~\ref{sec:results}. The record
length was selected on the validation split for each training-data
setting separately, the selected configuration was retrained on the
union of the training and validation sets, and the test split was
evaluated once; all numbers are means over three seeds. For fine-tuning,
the final SPF-only model initialises a training run on the OKC training
events alone with a ten times smaller learning rate ($10^{-4}$) for 15
epochs. Test metrics are reported separately for the OKC and the SPF
part of the test split.

\subsection{Results}

Table~\ref{tab:transfer-strategies} compares the four strategies. The
model trained on the SPF archive alone does not transfer to the modern
station: its macro-F1 on the OKC events is $0.39$, close to the value of
a classifier that assigns every event to the majority class ($0.31$),
although the same model reaches $0.96$ on the SPF part of the test
split. The confusion matrix (Table~\ref{tab:transfer-confusion}) shows
that the failure is specific. The archive model recognises 39 of the 46
OKC natural earthquakes and most of the blasts, but it labels about two
thirds of the modern mining-induced events as tectonic. The OKC
mining-induced events thus resemble the tectonic events of the 1990s
more than the mining-induced events of that time, as seen through the
archive model. Classifying all 17,928 OKC mining-induced events with the
archive model and relating the result to the bulletin metadata shows
that this is not a property of the events: the share labelled as
tectonic is the same for the events from the Czech (66\,\%) and the
Polish (73\,\%) part of the basin, it shows no trend over the years
2010--2023, and 28\,\% of the quarry blasts are labelled as tectonic as
well. It is a property of the records. The z-score-normalized OKC
records of all three classes are dominated by energy below 1\,Hz ---
mainly microseismic noise, which the broadband CMG-3ESP sensor passes
but the 2-Hz geophones of the SPF network suppressed --- so that their
spectral centroid (2--3\,Hz) lies below that of the archive tectonic
events (3.5\,Hz), and the P onset is barely visible above the
long-period background. In the archive, low-frequency, low-amplitude
records were the tectonic ones, and the archive model applies this rule
to the modern records. We verify this interpretation in
Section~\ref{sec:bandlimit}.

Training on the OKC data alone gives a macro-F1 of $0.915 \pm 0.005$ on
the OKC events. Blasts and mining-induced events are classified almost
perfectly (F1 of $0.98$ and $1.00$); the natural earthquakes, of which
only 107 are available for training, reach an F1 of $0.77$ (32 of the 46
test events recognised by the final model). Training on both data sets together gives
practically the same result on the OKC events ($0.918 \pm 0.007$, 32 of
46 tectonic events) while retaining the accuracy on the archive
($0.961$ on the SPF part). The archive data thus neither help nor harm
the classification of the modern records; a single model covers both
generations of instruments at no cost. Fine-tuning the archive model on
the OKC data is the worst of the three strategies that use OKC data
($0.864 \pm 0.007$) and, as expected, destroys the performance on the
archive: with the short schedule the network does not fully recover
from the archive-specific decision boundaries, whereas a network started
from ImageNet weights adapts to the OKC data directly.

Table~\ref{tab:transfer-length} separates the effect of the record
length from that of the training data on the validation split.
Extending the window from 8 to 16\,s improves the OKC macro-F1 from
$0.84$ to $0.90$ for the OKC-only model and from $0.87$ to $0.89$ for
the pooled model, mainly through the tectonic class, whose later phases
and coda fall outside the first eight seconds; a further extension to
20\,s brings no gain for the OKC-only model and a small one ($0.91$)
for the pooled model. At every record length the pooled and the
OKC-only model recognise the same number of the 21 tectonic validation
events within one or two events. The limiting factor for the natural earthquakes on the modern
station is therefore the number of labelled examples, not the model or
the amount of archive data: the archive earthquakes, recorded with
different instruments, do not substitute for examples from the station
itself.

For the binary tasks on the pooled data the picture is the same as on
the archive. The separation of blasts from mining-induced events reaches
a macro-F1 of $0.981$ ($0.986$ on the OKC events, with an area under the ROC curve of $0.999$), and
tectonic versus mining-induced events $0.964$ ($0.909$ on the OKC
events, limited again by the 46 tectonic test events).

\subsection{Band-limiting the records}
\label{sec:bandlimit}

If the low-frequency content of the modern records is the obstacle, a
common band limitation of both data sets should remove it. We therefore
repeated the experiment with a zero-phase Butterworth high-pass filter
(fourth order, cut-off 0.5 or 1\,Hz) applied to every record before the
normalization, leaving everything else unchanged
(Table~\ref{tab:bandlimit}). The filter is practically neutral for the
archive --- its records contain little energy below 1\,Hz to begin with
--- and the macro-F1 on the SPF part of the test split stays at
$0.96$ for both the archive-only and the pooled model, with the recall
of the archive tectonic events unchanged at $0.93$. On the OKC records
the archive model now recognises 87\,\% instead of 30\,\% of the
mining-induced events, which confirms the explanation given above.
Its transfer nevertheless remains unusable ($0.61$ instead of $0.39$):
with the low-frequency cue removed it still labels about 600 of the
5,379 mining-induced test events and a third of the blasts as tectonic,
so that the precision of the tectonic class is a few percent. The
remaining difference is in the natural earthquakes themselves --- the
archive contains several hundred records of the local earthquakes of
the 1990s, whereas fewer than twenty of the OKC earthquakes come from
the Ostrava area and the rest from other parts of the region and its
surroundings (Section~\ref{sec:okc-data}) --- and cannot be removed by
filtering. For the models that see OKC data the filter brings a small
but consistent gain: $0.931 \pm 0.005$ instead of $0.918 \pm 0.007$ on
the OKC events for the pooled model and $0.929 \pm 0.005$ instead of
$0.915 \pm 0.005$ for the OKC-only model, in both cases through the
tectonic class. (The filtered OKC-only model was selected at 20\,s
rather than 16\,s; since the unfiltered 20-s model is not better than
the 16-s one on the validation split, Table~\ref{tab:transfer-length},
the gain is due to the filter.) The two cut-off frequencies give the
same results within the seed variation.

\begin{table}[tbp]
  \centering
  \caption{Effect of a high-pass filter (0.5\,Hz) applied to both data
  sets before normalization: macro-F1 of the final EfficientNet-B0
  models ($D = 20$\,s) on the SPF and on the OKC part of the test
  split, mean $\pm$ std over three seeds, and the number of the 5,379 OKC
  mining-induced test events labelled as tectonic. The unfiltered
  OKC-only model is the 16-s model of Table~\ref{tab:transfer-strategies}.}
  \label{tab:bandlimit}
  \begin{tabular}{llccc}
    \toprule
    Training data & Filter & SPF test & OKC test & M $\to$ T (OKC) \\
    \midrule
    SPF only   & none   & $0.958 \pm 0.001$ & $0.387 \pm 0.006$ & 3,568 \\
    SPF only   & 0.5 Hz & $0.960 \pm 0.001$ & $0.607 \pm 0.008$ & 573 \\
    OKC only   & none   & --                & $0.915 \pm 0.005$ & 4 \\
    OKC only   & 0.5 Hz & --                & $0.929 \pm 0.005$ & 1 \\
    SPF + OKC  & none   & $0.961 \pm 0.001$ & $0.918 \pm 0.007$ & 2 \\
    SPF + OKC  & 0.5 Hz & $0.962 \pm 0.000$ & $0.931 \pm 0.005$ & 2 \\
    \bottomrule
  \end{tabular}
\end{table}

\section{Discussion and conclusions}
\label{sec:discussion}

The starting point of this work was a practical question of the
seismological service in the Ostrava region: can the three types of
events recorded there --- mining-induced events, natural earthquakes and
quarry blasts --- be told apart automatically from a short window around
the P-wave onset? The archive of the Fren\v{s}t\'{a}t network, once
recovered from its original format and cleaned, answers this question
for the records of the 1990s. The best model separates the three classes
with a macro-averaged F1 score of 0.96; blasts and natural earthquakes
are practically never confused with each other, and the remaining errors
consist of a few percent of the two rarer classes being absorbed by the
dominant mining-induced class. Since the SPF stations, the OKC station
and the sources of all three event types lie in the same area, this is
a strong indication that an operational ternary classifier for the
region is within reach.

Two methodological observations may be useful beyond this region. First,
the choice of the input representation mattered more than the choice of
the network. All models that saw the time--frequency structure of the
record --- the convolutional network and the hybrid transformer on
spectrograms --- clearly outperformed the recurrent models, which saw
either the magnitude spectrum of the whole window or the raw waveform,
and neither attention on the spectrum nor a parallel convolutional
branch on the waveform brought them closer. Longer
windows helped the spectrogram models and hurt the spectrum-based ones,
which points to the arrival of the S phase and the coda as the carriers
of the discriminating information. Second, the usual remedies for class
imbalance, loss weighting and undersampling, consistently lowered the
macro-F1 score: they trade a modest gain in recall of the rare classes
for a loss of precision that, with a 90\,\% majority class, dominates.
Training on the natural class distribution and judging the models by a
class-balanced metric was the better choice, and dedicated binary
classifiers improved on the ternary model by only about 0.01 for the
pairs of classes that involve the mining-induced events.

The transfer experiments qualify these results for the present-day
monitoring. A model trained on the archive alone is of no use on the
OKC station, although the two data sets come from the same area and are
normalized in the same way: the modern mining-induced events are mostly
labelled as tectonic by the archive model. The cause is not a change in the seismicity but in the recording: the
broadband sensor of the modern station passes the microseismic noise
below 1\,Hz that the geophones of the archive network suppressed, and
the archive model, for which low-frequency records were tectonic,
transfers this rule. A common high-pass filter removes the obstacle
without affecting the archive results and slightly improves the models
trained on the modern data (Section~\ref{sec:bandlimit}); what it
cannot remove is the different composition of the natural earthquakes in
the two catalogues --- the local earthquakes of the 1990s are well
represented in the archive and nearly absent from the modern data set
--- which is a matter of data rather than of modelling. Several limitations should also be kept in mind. The modern
data used in this study come from a single station and their tectonic class is limited to
the events of the bulletins of the Institute of Geonics; only 46 natural
earthquakes are available for testing, so the corresponding scores carry
an uncertainty of several hundredths, and the complete catalogue of the
Czech Regional Seismic Network remains to be exploited. The labels of both data sets are
those of the routine bulletins and have not been independently
re-examined. Both data sets consist of catalogued events only; the
models have never seen noise or non-seismic signals, and an operational
system therefore needs a detection and location stage before the
classifier. Finally,
the windows are anchored to P arrivals that were picked manually for the
OKC data and for part of the archive, and by PhaseNet for the rest of
the archive; in operation the anchor has to come from an automatic
picker, whose errors would propagate to the classifier and have not
been evaluated end to end.

For the present network the results suggest a pragmatic path. Blasts and
mining-induced events can already be separated automatically on the OKC
station from a 16-s window with F1 scores above 0.97, and natural
earthquakes are recognised with a recall of 0.7--0.8, which is
sufficient to flag candidates for manual review but not yet for
unattended classification. Training on the OKC data alone and training
on the OKC data together with the archive give the same accuracy on the
modern records, so the archive can be included without penalty, and a
single model then covers both generations of instruments. The decisive
improvement, however, will come from enlarging the set of catalogued
natural earthquakes on the modern stations, not from further modelling;
the archive earthquakes, recorded with different instruments, do not
substitute for them.

The natural continuation of this work is therefore the extension to the
other stations of the Czech Regional Seismic Network in the region,
which multiplies the number of modern records of all three classes and,
in particular, of natural earthquakes, together with a detection and
location stage and a scheme for retraining the classifier as the
catalogue grows. In such a system the classification follows the
location of the event, so that the location --- local or distant,
inside or outside the mining area --- can be taken into account, for
example by restricting the classifier to local events or by training
separate classifiers for different distance ranges. The ultimate aim is
to support the automated processing of the whole Czech Regional Seismic
Network. The
data preparation pipeline and the classifiers presented here are
designed to make this step straightforward. Neither is tied to the
region: the cleaning pipeline operates on the waveforms alone, and the
classifiers can be retrained on the records of any station network with
a labelled catalogue. The value of the Ostrava region for such a study
lies in the coexistence of all three event types within one small area,
which makes the ternary problem both necessary and testable there.

\section*{Acknowledgements}
The authors thank Bohuslav R\r{u}\v{z}ek for valuable advice about deterministic methods of data cleaning.

This work was supported by the programme Dynamic Planet Earth of the Czech Academy of Sciences -- Strategy AV21.

Michael Skotnica was supported by Programme to Support Prospective Human
Resources -- Postdoctoral Fellows of the Czech Academy of Sciences, reg. no. L100862601.

\section*{Use of generative AI}
The language of the paper was polished with the help of Claude
(Anthropic), which also served as a coding assistant in writing and
debugging the analysis scripts. The authors checked and revised all
text and code produced in this way and are fully responsible for the
content of the paper and for the reported analyses.

\section*{Data and code availability}
The data-cleaning pipeline for the SPF archive (format conversion,
quality checks and P-wave picking) is available at
\url{https://github.com/natural-hazards/SPF_cleaning}, and the
classification models and training pipelines at
\url{https://github.com/natural-hazards/earthquakes-nn}. The SPF and OKC
event data sets used in this study are available at Zenodo,
\url{https://doi.org/10.5281/zenodo.24010919}.

\bibliographystyle{alpha}
\bibliography{bibliography}
\end{document}